\PassOptionsToPackage{unicode}{hyperref}
\PassOptionsToPackage{hyphens}{url}
\documentclass[
  11pt,
  letterpaper,
]{article}
\usepackage{xcolor}
\usepackage[margin=1in,headheight=15pt,headsep=0.24in,footskip=0.38in]{geometry}
\usepackage{amsmath,amssymb}
\usepackage{iftex}
\ifPDFTeX
  \usepackage[T1]{fontenc}
  \usepackage[utf8]{inputenc}
  \usepackage{textcomp}
  \IfFileExists{newpxtext.sty}{%
    \usepackage{newpxtext}
    \usepackage{newpxmath}
  }{%
    \usepackage{lmodern}
  }
\else
  \usepackage{unicode-math}
  \defaultfontfeatures{Scale=MatchLowercase}
  \defaultfontfeatures[\rmfamily]{Ligatures=TeX,Scale=1}
\fi
\ifPDFTeX\else
\fi
\IfFileExists{upquote.sty}{\usepackage{upquote}}{}
\IfFileExists{microtype.sty}{% use microtype if available
  \usepackage[]{microtype}
  \UseMicrotypeSet[protrusion]{basicmath} % disable protrusion for tt fonts
}{}
\makeatletter
\@ifundefined{KOMAClassName}{% if non-KOMA class
  \IfFileExists{parskip.sty}{%
    \usepackage{parskip}
  }{% else
    \setlength{\parindent}{0pt}
    \setlength{\parskip}{6pt plus 2pt minus 1pt}}
}{% if KOMA class
  \KOMAoptions{parskip=half}}
\makeatother
\usepackage{color}
\usepackage{fancyvrb}

\DefineVerbatimEnvironment{Highlighting}{Verbatim}{commandchars=\\\{\}}
\newenvironment{Shaded}{}{}

\newcommand{\BuiltInTok}[1]{\textcolor[rgb]{0.00,0.50,0.00}{#1}}

\newcommand{\KeywordTok}[1]{\textcolor[rgb]{0.00,0.44,0.13}{\textbf{#1}}}
\newcommand{\NormalTok}[1]{#1}

\newcommand{\StringTok}[1]{\textcolor[rgb]{0.25,0.44,0.63}{#1}}

\usepackage{longtable,booktabs,array}
\usepackage{calc} % for calculating minipage widths
\usepackage{etoolbox}
\makeatletter
\patchcmd\longtable{\par}{\if@noskipsec\mbox{}\fi\par}{}{}
\makeatother
\IfFileExists{footnotehyper.sty}{\usepackage{footnotehyper}}{\usepackage{footnote}}
\makesavenoteenv{longtable}
\usepackage{graphicx}
\makeatletter
\newsavebox\pandoc@box
\newcommand*\pandocbounded[1]{% scales image to fit in text height/width
  \sbox\pandoc@box{#1}%
  \Gscale@div\@tempa{\textheight}{\dimexpr\ht\pandoc@box+\dp\pandoc@box\relax}%
  \Gscale@div\@tempb{\linewidth}{\wd\pandoc@box}%
  \ifdim\@tempb\p@<\@tempa\p@\let\@tempa\@tempb\fi% select the smaller of both
  \ifdim\@tempa\p@<\p@\scalebox{\@tempa}{\usebox\pandoc@box}%
  \else\usebox{\pandoc@box}%
  \fi%
}
\def\fps@figure{htbp}
\makeatother
\NewDocumentCommand\citeproctext{}{}
\NewDocumentCommand\citeproc{mm}{%
  \begingroup\def\citeproctext{#2}\cite{#1}\endgroup}
\makeatletter
 \let\@cite@ofmt\@firstofone
 \def\@biblabel#1{}
 \def\@cite#1#2{{#1\if@tempswa , #2\fi}}
\makeatother
\newlength{\cslhangindent}
\newlength{\csllabelwidth}
\newenvironment{CSLReferences}[2] % #1 hanging-indent, #2 entry-spacing
 {\begin{list}{}{%
  \setlength{\itemindent}{0pt}
  \setlength{\leftmargin}{0pt}
  \setlength{\parsep}{0pt}
  \ifodd #1
   \setlength{\leftmargin}{\cslhangindent}
   \setlength{\itemindent}{-1\cslhangindent}
  \fi
  \setlength{\itemsep}{#2\baselineskip}}}
 {\end{list}}
\usepackage{calc}

\providecommand{\tightlist}{%
  \setlength{\itemsep}{0pt}\setlength{\parskip}{0pt}}
\definecolor{ArticleGray}{gray}{0.36}
\definecolor{CaptionGray}{gray}{0.45}
\IfFileExists{titlesec.sty}{%
  \usepackage{titlesec}
  \titleformat{\section}{\large\sffamily\bfseries}{\thesection}{0.7em}{}
  \titleformat{\subsection}{\normalsize\sffamily\bfseries}{\thesubsection}{0.7em}{}
  \titleformat{\subsubsection}{\normalsize\sffamily\itshape}{\thesubsubsection}{0.7em}{}
  \titlespacing*{\section}{0pt}{1.35\baselineskip}{0.55\baselineskip}
  \titlespacing*{\subsection}{0pt}{1.1\baselineskip}{0.4\baselineskip}
  \titlespacing*{\subsubsection}{0pt}{0.95\baselineskip}{0.3\baselineskip}
}{}
\usepackage{graphicx}
\IfFileExists{tikz.sty}{%
  \usepackage{tikz}
  \usetikzlibrary{arrows.meta,positioning,calc,fit,shapes.geometric}
}{}
\usepackage{etoolbox}
\makeatletter

\newcommand{\ArticleCaption}[1]{%
  \par\smallskip
  {\small\itshape\color{CaptionGray}#1\par}%
  \smallskip
}
\long\def\@makecaption#1#2{%
  \vskip\abovecaptionskip
  \sbox\@tempboxa{{\small\itshape\color{CaptionGray}\textbf{#1.} #2}}%
  \ifdim \wd\@tempboxa >\hsize
    {\small\itshape\color{CaptionGray}\textbf{#1.} #2\par}%
  \else
    \global \@minipagefalse
    \hb@xt@\hsize{\hfil\box\@tempboxa\hfil}%
  \fi
  \vskip\belowcaptionskip
}
\def\ps@articlemodern{%
  \def\@oddhead{\vbox{%
    \hbox to\textwidth{\footnotesize\color{CaptionGray}\sffamily Workflow
as Knowledge: Semantic Persistence for LLM-Mediated
Workflows\hfil }%
    \vskip 3pt%
    {\color{CaptionGray}\hrule height 0.35pt}%
  }}%
  \def\@evenhead{\@oddhead}%
  \def\@oddfoot{\hfil{\footnotesize\color{ArticleGray}\sffamily \thepage}\hfil}%
  \def\@evenfoot{\@oddfoot}%
}
\makeatother
\AtBeginDocument{\pagestyle{articlemodern}}
\usepackage{bookmark}
\IfFileExists{xurl.sty}{\usepackage{xurl}}{} % add URL line breaks if available
\hypersetup{
  pdftitle={Workflow as Knowledge: Semantic Persistence for LLM-Mediated Workflows},
  pdfauthor={Emanuele Quinto; Carlo Andrea Rozzi; Francesco Zanitti},
  hidelinks,
  pdfcreator={LaTeX via pandoc}}

\title{\sffamily\bfseries Workflow as Knowledge: Semantic Persistence
for LLM-Mediated Workflows}
\author{%
\begin{minipage}[t]{0.30\textwidth}
\centering
\small
\textbf{Emanuele Quinto}\\
UNHCR\\
København, Denmark\\
\texttt{emanuele.quinto@protonmail.ch}
\end{minipage}
\hfill%
\begin{minipage}[t]{0.30\textwidth}
\centering
\small
\textbf{Carlo Andrea Rozzi}\\
CNR—Istituto Nanoscienze\\
Modena, Italy
\end{minipage}
\hfill%
\begin{minipage}[t]{0.30\textwidth}
\centering
\small
\textbf{Francesco Zanitti}\\
ZeLe \& F ApS\\
København, Denmark
\end{minipage}
}

\date{}

\begin{document}
\maketitle
\thispagestyle{articlemodern}

\section{Abstract}\label{abstract}

Large language model (LLM) applications increasingly use explicit
workflows for tool use, retrieval, branching, checkpointing, and human
approval. Existing workflow systems already address many execution
concerns. This paper proposes a Lisp-inspired but language-independent
conceptual model: symbolic forms, object identity, and live-image
thinking are used as explanatory lenses, not implementation commitments.
In this model, workflow definitions, workflow instances, inference
records, context snapshots, and dependency relations are represented as
persistent knowledge objects in a shared knowledge substrate. Its
central semantic distinction is between \texttt{derive} and
\texttt{infer}: \texttt{derive} is deterministic computation over
available state; \texttt{infer} is mediated LLM judgment under declared
context and executor-controlled capability policy.

The result is a preliminary conceptual account of semantic persistence:
workflows do not merely produce knowledge and leave traces, but can
themselves be represented as inspectable, resumable, and reviewable
knowledge objects, while formal transition semantics remain future work.

\section{1. Introduction}\label{introduction}

LLM systems have moved from single-turn prompting toward structured
workflows. Agent frameworks now expose graphs, typed steps, loops, tool
calls, checkpointing, human approval gates, and reusable submodules
(\citeproc{ref-agentspex2026}{\emph{{AgentSPEX}} 2026};
\citeproc{ref-langgraphPersistence2026}{LangGraph 2026};
\citeproc{ref-khattab2023dspy}{{Khattab et al.} 2023};
\citeproc{ref-josifoski2023flows}{{Josifoski et al.} 2023};
\citeproc{ref-workflowllm2024}{\emph{{WorkflowLLM}} 2024}). This shift
is necessary: unconstrained reactive prompting leaves control flow
implicit, makes intermediate state difficult to inspect, and complicates
resumption after interruption.

However, making execution explicit does not by itself solve a deeper
representational problem. A workflow definition may live as source code,
declarative configuration, or a database-backed representation. A
running occurrence is managed by a runtime layer. As of 2026,
practitioner discussions often describe parts of this surrounding
control-and-runtime machinery as an \texttt{agent\ harness}
(\citeproc{ref-oreilly2026agentHarness}{O'Reilly 2026a}). Intermediate
model outputs may be retained as logs, traces, rows, or chat history.
Such artifacts can be correlated by infrastructure or provenance
mechanisms, but correlation is not the same as assigning them explicit
roles in the shared semantic object model developed here.

Our proposal first separates three conceptual layers. A lower
\textbf{runtime service layer} supplies model adapters, tools, external
processes, and persistence/indexing facilities.

A middle \textbf{control layer} contains the domain-specific language
(\texttt{DSL}) machine and its executor: it interprets declared objects,
assembles context, mediates model and tool calls, validates results,
applies permitted transitions, and exposes the knowledge-substrate
interface.

A higher \textbf{semantic layer} contains workflow definitions, workflow
instances, and their linked inference, approval, and panel records.
Contemporary \texttt{agent\ harness} implementations may package
portions of the middle and lower layers together; that packaging is not
itself the definition of a workflow.

Figure 1 summarizes the three conceptual layers and the mediation
boundary between semantic objects, the DSL-machine control layer, and
runtime services.

\begin{figure}
\centering
\pandocbounded{\includegraphics[width=0.82\linewidth,height=0.7\textheight,keepaspectratio,alt={Figure 1. Semantic workflow objects are interpreted by the DSL-machine control layer, which coordinates runtime services and writes back workflow instances, mediated effects, and records of inference, approval, and panel activity. The bidirectional relation indicates that the control layer both reads semantic objects and writes back persistent semantic objects or relations.}]{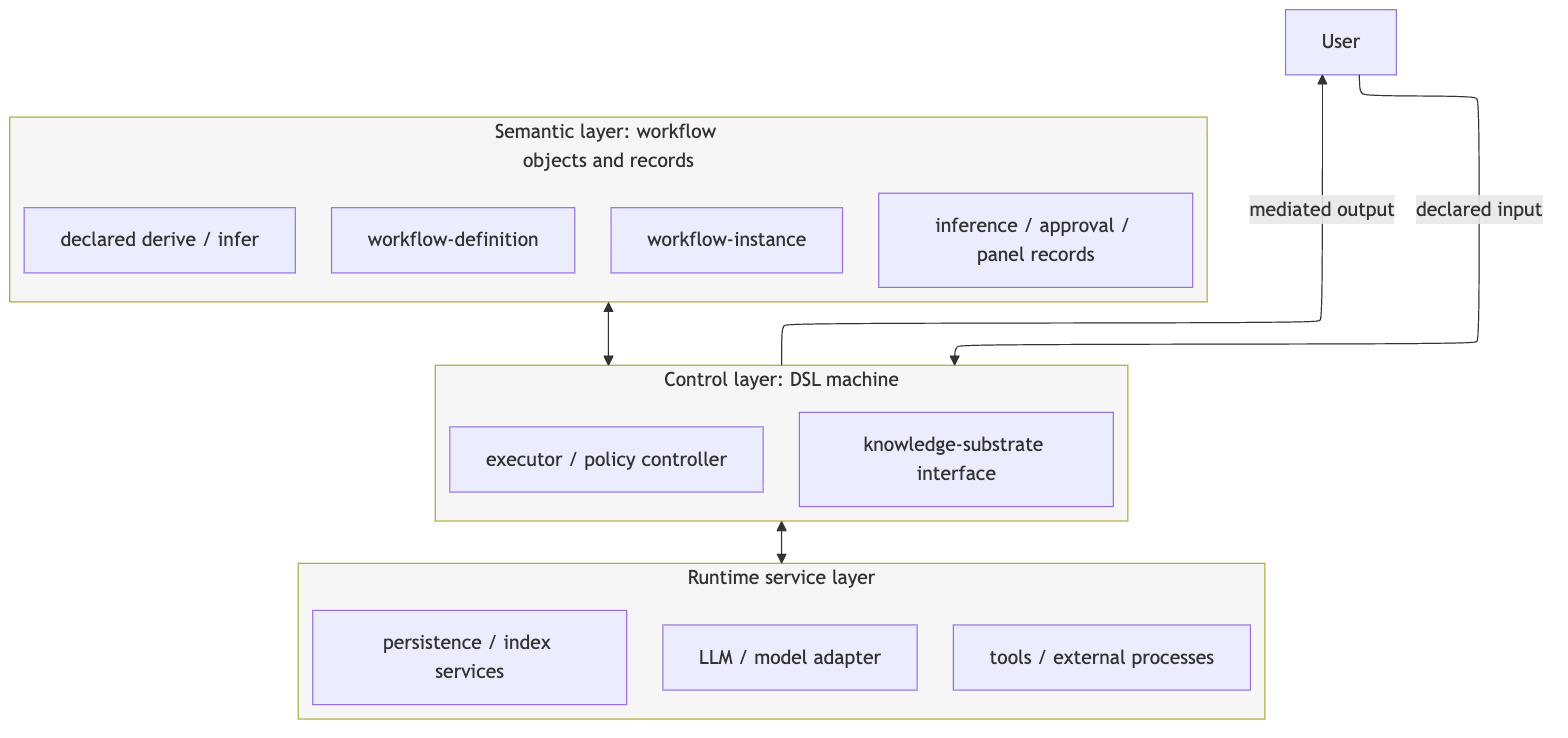}}
\caption{Semantic workflow objects are interpreted by the
DSL-machine control layer, which coordinates runtime services and writes
back workflow instances, mediated effects, and records of inference,
approval, and panel activity. The bidirectional relation indicates that
the control layer both reads semantic objects and writes back persistent
semantic objects or relations.}\label{fig:mermaid-001}
\end{figure}

Within this architecture, a \texttt{workflow-definition} is a semantic
object that specifies a declared process of steps, state transitions,
and human or model-mediated interactions.

When the executor interprets it, a \texttt{workflow-instance} is created
or resumed. Its consequential inference, approval, and deliberation
records are linked as distinct semantic objects.

This paper proposes that workflow definitions, workflow instances, and
linked inference records persist as queryable knowledge objects within a
shared knowledge substrate. Lisp supplies the explanatory lens: symbolic
forms that stay close to the object structures they denote, object
identity, reflective inspection, interactive live-image development, and
extensible object protocols provide a natural way to state the model.

The proposal is conceptual; implementation and empirical evaluation are
left for future work. In this paper, \textbf{semantic persistence} means
preserving typed workflow objects and inference records, their
identities and relations, and the context needed for later inspection
beyond a single execution run. It does not by itself imply accurate
attribution, reproducibility, trust, or audit quality.

\textbf{Workflow as knowledge} names the shift from treating workflows
only as executable control structures to treating their definitions,
running instances, and consequential inference records as persistent
semantic objects.

The model was abstracted from broader work on inspectable
personal-computing environments in which users should be able to review
the workflows, model-mediated judgments, and records that shape later
action. That design context motivates the emphasis on object identity,
mediated authority, and durable inference records, but the account here
is deliberately generalized and does not depend on a particular device
architecture or implementation.

The paper makes four claims:

\begin{enumerate}
\def\labelenumi{\arabic{enumi}.}
\tightlist
\item
  Workflow definitions should be represented as data objects, not only
  as executable procedures or external configuration.
\item
  Workflow instances should persist as live objects whose state can
  serve as a checkpoint.
\item
  LLM calls should be syntactically and semantically distinguished from
  deterministic state computations through \texttt{infer} and
  \texttt{derive}.
\item
  Human-facing reasoning interactions should persist as typed records,
  not disappear as UI events or unstructured chat turns, a framing
  motivated by HITL role distinctions and decision-provenance concerns
  (\citeproc{ref-mosqueiraRey2023hitl}{Mosqueira-Rey et al. 2023};
  \citeproc{ref-singh2019decisionProvenance}{Singh et al. 2019}).
\end{enumerate}

This paper is a conceptual model proposal, not an empirical study or a
formal calculus. Its guiding question is whether workflow definitions,
running instances, and the inference records they produce can be
represented as typed, persistent knowledge objects in a shared
substrate, while deterministic computation and LLM-mediated judgment are
held to distinct semantic standards. The contribution is the object
model and vocabulary itself: the semantic object schema, the primitive
vocabulary, the \texttt{derive} / \texttt{infer} boundary, and the
preliminary provenance comparison in Appendix C. The four claims above
are design commitments that follow from this framing; Section 6
identifies which of them require future empirical or formal evaluation
before they could be treated as tested claims.

\section{2. Related Work}\label{related-work}

This section reviews knowledge-management and execution systems that
address related representational problems: how processes are described,
how runs are retained, how inference or provenance can be inspected
later, and how human or model-mediated decisions are situated in a
larger system. The comparison is selective. It identifies adjacent
mechanisms and boundaries rather than claiming that these systems solve
the same problem as the proposed object model. The review moves from
provenance, notebook history, hypertext, and symbolic environments
toward LLM-specific DSL generation, human-in-the-loop systems, agent
workflow systems, and practitioner accounts of agent harnesses.

\subsection{2.1 Provenance}\label{provenance}

Provenance research makes the boundary between execution persistence and
semantic persistence more demanding. Scientific workflow systems already
represent workflow specifications, execution histories, and derived data
so that prior runs can be queried, compared, reused, or analyzed for
reproducibility
(\citeproc{ref-davidson2008scientificProvenance}{Davidson and Freire
2008}). Control-flow provenance work further argues that branches and
loops must be represented when execution structure matters to
understanding a result
(\citeproc{ref-butt2021controlFlowProvenance}{Butt and Fitch 2021}).
Decision provenance and World Wide Web Consortium (\texttt{W3C}) PROV
provide vocabulary for linking entities, activities, agents, decisions,
and consequences (\citeproc{ref-singh2019decisionProvenance}{Singh et
al. 2019}; \citeproc{ref-w3c2013prov}{W3C 2013}).

Recent agent-provenance work extends this concern to prompts, responses,
decisions, workflow context, and downstream effects
(\citeproc{ref-souza2025provAgent}{{Souza et al.} 2025b}). Other work
uses LLM agents to query scientific-workflow provenance directly
(\citeproc{ref-souza2025interactiveWorkflowProvenance}{{Souza et al.}
2025a}).

The contribution here is a typed, Lisp-inspired semantic model for
workflow definitions, inference dependencies, approvals, and structured
human deliberations as persistent knowledge objects.

\subsection{2.2 Hypertext}\label{hypertext}

Computational notebooks and hypertext systems provide two additional
adjacent precedents. ProvBook represents Jupyter cell executions and
results as queryable provenance, and Jupyter's Archive records navigable
output histories from exploratory notebook work
(\citeproc{ref-samuel2018provbook}{Samuel and König-Ries 2018};
\citeproc{ref-chaudhary2019jupyterArchive}{Chaudhary 2019}). Earlier
hypertext work addresses durable, richly connected information
structures and formal networks of components, links, anchors, and
interaction layers (\citeproc{ref-nelson1965fileStructure}{Nelson 1965};
\citeproc{ref-halasz1994dexter}{Halasz and Schwartz 1994}). These
traditions show that retained execution history and linked knowledge
structures are not new ideas; the present model applies similar
persistence pressure to typed LLM inference, approval, and deliberation
records within an executable workflow model.

\subsection{2.3 Lisp And Symbolic
Environments}\label{lisp-and-symbolic-environments}

Other work explores Lisp and symbolic environments for LLMs. De la Torre
proposes integrating LLMs with a persistent Lisp metaprogramming loop,
where generated Lisp expressions are intercepted and evaluated in a live
environment (\citeproc{ref-delatorre2025lispLoop}{Torre 2025}). Pel
explores a Lisp-flavored language for LLM action expression and
orchestration (\citeproc{ref-mohammadi2025pel}{Mohammadi 2025}). These
systems motivate the relevance of homoiconicity and persistent symbolic
state, but they generally give the LLM a more direct role in generating
or evolving executable expressions.

\subsection{2.4 LLM-Assisted DSL
Generation}\label{llm-assisted-dsl-generation}

Work on LLM-assisted DSL generation is relevant for a narrower reason.
DSL-Xpert and its journal extension show that LLMs can help translate
natural language into DSL expressions when supplied with grammars,
examples, validation, and correction loops
(\citeproc{ref-dslxpert2024}{\emph{{DSL-Xpert}} 2024};
\citeproc{ref-dslxpert2025}{{``{DSL-Xpert} 2.0''} 2025};
\citeproc{ref-dslGenerationFinetuningRag2024}{\emph{A Comparative Study
of {DSL} Code Generation} 2024}). Cabot's experience report on LLMs for
DSL development is similarly cautious: LLMs may assist language
engineers, but they still require guidance and risk-sensitive tool
support (\citeproc{ref-cabot2026llmsDslDevelopment}{Cabot 2026}). These
sources support using LLMs as assistants for candidate DSL artifacts,
not as autonomous language authorities.

The design proposed here is more conservative about authority. The LLM
cannot directly introduce workflow transitions or execute external
actions: it fills declared \texttt{infer} slots under executor rules.
Its inferred values may nevertheless influence a declared branch applied
by the executor. Those dependencies should be explicit records rather
than hidden causal steps. In this sense the LLM is a mediated
peripheral, not an autonomous process supervisor.

\subsection{2.5 Human-In-The-Loop
Systems}\label{human-in-the-loop-systems}

Human-in-the-loop frameworks, including aiFlows and modern agent
checkpointing systems, also inform this proposal
(\citeproc{ref-josifoski2023flows}{{Josifoski et al.} 2023};
\citeproc{ref-langgraphPersistence2026}{LangGraph 2026}). HITL
literature distinguishes different allocations of human and machine role
and authority rather than a single generic ``human step''
(\citeproc{ref-mosqueiraRey2023hitl}{Mosqueira-Rey et al. 2023}). The
difference proposed here is representational: an approval is an
authorization event, while a panel is a structured deliberation
occurrence containing its motion, arguments, decision, and context.

\subsection{2.6 Agent Workflow Systems}\label{agent-workflow-systems}

Agent workflow systems such as AgentSPEX, LangGraph, DSPy,
Flows/aiFlows, WorkflowLLM, and related orchestration frameworks address
the execution control problem from several directions. AgentSPEX is
especially close in motivation, offering a declarative workflow language
with typed steps, explicit state management, checkpointing,
verification, and logging
(\citeproc{ref-agentspex2026}{\emph{{AgentSPEX}} 2026}). LangGraph
models agent behavior as stateful graphs with checkpoint-based
persistence for resumption, human inspection, memory, time travel, and
fault tolerance (\citeproc{ref-langgraphPersistence2026}{LangGraph
2026}). DSPy treats LLM applications as declarative modules that can be
optimized into pipelines (\citeproc{ref-khattab2023dspy}{{Khattab et
al.} 2023}). Flows and aiFlows model AI-AI and human-AI interactions as
composable blocks with isolated state and message-based interfaces
(\citeproc{ref-josifoski2023flows}{{Josifoski et al.} 2023}).
WorkflowLLM represents a data-centric approach to improving workflow
orchestration over many application programming interfaces
(\texttt{APIs}) (\citeproc{ref-workflowllm2024}{\emph{{WorkflowLLM}}
2024}).

These systems show that explicit structure, state, checkpointing,
modularity, and orchestration are already active concerns in LLM
application engineering. The present proposal differs in its target
abstraction. It does not claim that existing systems lack persistence.
Rather, it distinguishes execution persistence from semantic
persistence: the question is whether workflow definitions, workflow
instances, and inference records should remain as first-class knowledge
objects after execution.

\begin{figure}
\centering
\pandocbounded{\includegraphics[width=0.82\linewidth,height=0.7\textheight,keepaspectratio,alt={Figure 2. Execution persistence retains runnable state, checkpoints, logs, traces, and outputs; semantic persistence treats workflow definitions, workflow instances, inference records, and context snapshots as first-class knowledge objects.}]{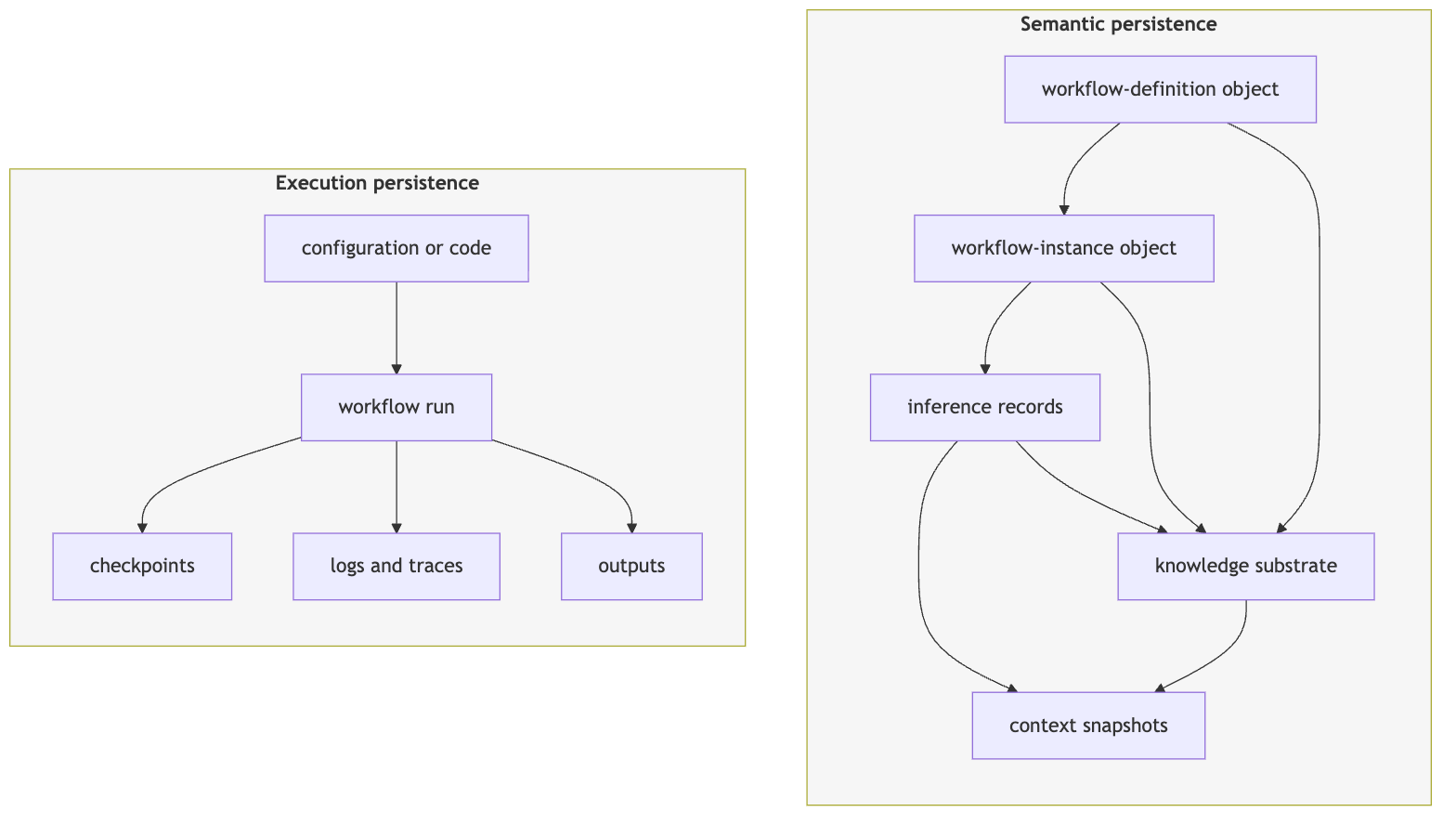}}
\caption{Execution persistence retains runnable state,
checkpoints, logs, traces, and outputs; semantic persistence treats
workflow definitions, workflow instances, inference records, and context
snapshots as first-class knowledge objects.}\label{fig:mermaid-002}
\end{figure}

\subsection{2.7 Agent Harnesses}\label{agent-harnesses}

Recent practitioner discussions of agent harnesses make the same
boundary visible from an engineering direction: the system around the
model shapes context, tool use, permissions, feedback, observability,
and recovery (\citeproc{ref-oreilly2026agentHarness}{O'Reilly 2026a}).
Practitioner essays on coding-agent harnesses add useful vocabulary for
feedforward guidance, feedback sensors, computational checks,
inferential reviews, permissions, and lifecycle hooks
(\citeproc{ref-humanlayer2026harnessEngineering}{HumanLayer 2026};
\citeproc{ref-fowler2026harnessEngineering}{Böckeler 2026}). This paper
treats harness engineering as current practitioner framing, not as a
settled academic field, and keeps the contribution at the level of
semantic workflow representation.

\section{3. The Core Abstraction: Workflow Objects In A Knowledge
Substrate}\label{the-core-abstraction-workflow-objects-in-a-knowledge-substrate}

\subsection{3.1 The Fragmentation
Problem}\label{the-fragmentation-problem}

Contemporary LLM-augmented workflows are typically implemented as
pipelines: a sequence of function calls, API invocations, retrieval
operations, and prompt templates connected by imperative glue code or
declarative configuration. The workflow definition exists as source code
or configuration. The running workflow exists as an execution thread.
The results - model outputs, user choices, and intermediate artifacts -
exist as logs, database rows, or conversation history.

Without a provenance or knowledge representation designed for such
queries, this fragmentation has practical consequences. A workflow that
ran six months ago may not be queryable alongside the knowledge it
produced. A structured deliberation that informed a decision may not be
traversable as a graph node linked to the decision record. A deferred
question may not automatically surface when related material appears
later. The proposal targets inference and deliberation acts that
otherwise remain structurally outside the knowledge objects they shape.

Recent agent workflow languages address part of this problem by making
control flow explicit. This is a real advance over reactive prompting.
The remaining question is whether the workflow, together with its
inference and deliberation acts, can be represented with the same
semantic status as the knowledge they help produce.

\subsection{3.2 Live Images As Analogy, Not
Precedent}\label{live-images-as-analogy-not-precedent}

Walker et al.~describe Symbolics Genera as a Lisp-based programming
environment integrating operating-system, utility, and application
facilities around shared in-memory objects, multiple processes, windows,
and development tools
(\citeproc{ref-walker1987symbolicsGeneraProgrammingEnvironment}{Walker
et al. 1987}). Archived Symbolics documentation adds the saved-world
terminology: a saved Genera world consists of Lisp objects and can be
inspected as part of that environment
(\citeproc{ref-symbolicsGeneraConcepts1990}{Symbolics, Inc. 1990}).

This paper is inspired by that live-image tradition because it shows a
computing style in which programs, data, tools, and user-visible state
can be inspected in one integrated environment. This paper does not
claim that Genera directly solved LLM workflow persistence. Instead, it
borrows the live-image intuition and applies it to a different target:
workflow definitions, running workflow instances, and inference
artifacts should inhabit the same semantic knowledge substrate as notes,
decisions, documents, or other knowledge objects.

The live-image analogy matters because LLM workflows are often
long-running, stateful, interrupted, resumed, and revised. Their
intermediate products are frequently as important as final outputs. A
decision summary is useful, but the structured reasoning that led to it
is also useful. In the proposed object model, that reasoning is not
merely an external trace. It is part of the knowledge substrate.

\subsection{3.3 A Core Semantic Schema}\label{a-core-semantic-schema}

Conceptually, the model introduces semantic object kinds and relations
for workflow definitions, workflow instances, inference records, context
snapshots, and dependency links. A Lisp-inspired notation is useful
because a written form can remain close to the object structure it
denotes, following Lisp's symbolic-expression tradition
(\citeproc{ref-mccarthy1960recursive}{McCarthy 1960}). A
\texttt{define-workflow} form therefore denotes a queryable
workflow-definition object containing declared inputs, resources, state
schema, guard rules, and steps, rather than only an opaque procedure or
external configuration.

In this paper, the \texttt{knowledge\ substrate} is an abstract semantic
persistence layer: it gives workflow definitions, workflow instances,
inference records, context snapshots, and dependency relations stable
identity, typed roles, typed relations, and queryability across
execution boundaries. It is not a commitment to a particular storage
technology such as a graph database, RDF store, object store, event log,
relational database, or Lisp image; those are possible implementations
of the interface, not the model itself.

A Common Lisp implementation could realize such objects using Common
Lisp Object System (\texttt{CLOS}) classes, instances, slots, and
generic functions, one possible path among several
(\citeproc{ref-bobrow1988closSpec}{Bobrow et al. 1988};
\citeproc{ref-gabriel1991clos}{Gabriel et al. 1991}). Appendix A
sketches the primitive effects, and Appendix B provides the illustrative
notation and worked example.

Table 1 names semantic objects that may persist in the knowledge
substrate. It is illustrative rather than exhaustive: the point is to
show the kinds of workflow structure, context, computation,
model-mediated output, human decision, and relation that can become
durable knowledge objects.

{\def\LTcaptype{none} % do not increment counter
\begin{longtable}[]{@{}
  >{\raggedright\arraybackslash}p{(\linewidth - 2\tabcolsep) * \real{0.4600}}
  >{\raggedright\arraybackslash}p{(\linewidth - 2\tabcolsep) * \real{0.4600}}@{}}
\toprule\noalign{}
\begin{minipage}[b]{\linewidth}\raggedright
Role
\end{minipage} & \begin{minipage}[b]{\linewidth}\raggedright
Semantic objects
\end{minipage} \\
\midrule\noalign{}
\endhead
\bottomrule\noalign{}
\endlastfoot
Workflow structure & \texttt{workflow-definition}: reusable declared
process; \texttt{workflow-instance}: one running or resumed
occurrence \\
Inputs, resources, and context & \texttt{input-binding}: declared user
or system input; \texttt{resource-binding}: resource made available
under workflow rules; \texttt{context-snapshot}: bounded material
visible for an inference, approval, or decision;
\texttt{retrieval-record}: retrieved material when retrieval matters to
later review \\
Computation and model-mediated results & \texttt{derived-object}:
deterministic result produced by \texttt{derive};
\texttt{inference-record}: mediated model judgment with declared
context; \texttt{tool-result-record}: result returned by a permitted
tool or external process \\
Human and policy records & \texttt{decision-record}: user or policy
choice; \texttt{approval-record}: human authorization or rejection gate;
\texttt{panel-record}: structured deliberation occurrence with motion,
options, arguments, decision, and context \\
Templates and relations & \texttt{panel-template}: reusable deliberation
structure; \texttt{dependency-link}: relation from an object to premises
or records it depends on; \texttt{supersession-link}: relation between
revised, disputed, or replaced objects \\
\end{longtable}
}

\textbf{Table 1.} Core semantic object kinds and relations in the
proposed model.

Table 2 names the operating vocabulary used to construct and interpret
these objects in the \texttt{DSL\ machine}. The first table says what
may persist as a semantic object or relation; the second says what a
workflow author can declare or invoke. Runtime mechanisms such as
retrieval-augmented generation (\texttt{RAG}), Model Context Protocol
(\texttt{MCP})-style connectors, tool APIs, or databases are not
semantic objects by themselves. They become visible in the model when
their inputs, outputs, retrieved context, decisions, or effects are
represented as typed objects or records.

{\def\LTcaptype{none} % do not increment counter
\begin{longtable}[]{@{}
  >{\raggedright\arraybackslash}p{(\linewidth - 4\tabcolsep) * \real{0.1964}}
  >{\raggedright\arraybackslash}p{(\linewidth - 4\tabcolsep) * \real{0.3721}}
  >{\raggedright\arraybackslash}p{(\linewidth - 4\tabcolsep) * \real{0.3515}}@{}}
\toprule\noalign{}
\begin{minipage}[b]{\linewidth}\raggedright
Primitive
\end{minipage} & \begin{minipage}[b]{\linewidth}\raggedright
Role in the model
\end{minipage} & \begin{minipage}[b]{\linewidth}\raggedright
Associated object or relation
\end{minipage} \\
\midrule\noalign{}
\endhead
\bottomrule\noalign{}
\endlastfoot
\texttt{resource} & Material loaded or bound into the workflow &
resource reference \\
\texttt{guard} & Constraint, permission rule, or scope lock & guard
result or violation \\
\texttt{context} & Information visible or authoritative for a step &
context snapshot \\
\texttt{state} & Active workflow state used during execution &
workflow-instance slot \\
\texttt{record} & Durable output or evidence after execution & typed
record \\
\texttt{approval} & Permission or confirmation gate & approval record \\
\texttt{panel} & Structured human deliberation & panel record \\
\texttt{derive} & Deterministic computation over state & derived
value \\
\texttt{infer} & Mediated LLM judgment under declared context &
inference record \\
\texttt{capability} /\allowbreak{} \texttt{action} & Side-effecting tools -
\emph{Candidate refinement} & capability policy or action record \\
\texttt{handoff} /\allowbreak{} \texttt{promotion} & Routing or elevating knowledge -
\emph{Candidate refinement} & handoff or promotion record \\
\end{longtable}
}

\textbf{Table 2.} Operating vocabulary for constructing and interpreting
workflow objects.

Two distinctions deserve early emphasis. First, \texttt{state} is active
workflow state used while execution continues, while \texttt{record} is
durable output or evidence retained after execution. Second, within the
human-facing records, an \texttt{approval} represents authority to
permit, reject, or defer a transition, while a \texttt{panel} represents
structured deliberation whose result may later inform an
executor-applied transition.

\subsection{\texorpdfstring{3.4 \texttt{derive}, \texttt{infer}, And
Context}{3.4 derive, infer, And Context}}\label{derive-infer-and-context}

The DSL distinguishes deterministic computation from LLM-mediated
judgment:

\begin{Shaded}
\begin{Highlighting}[]
\NormalTok{(derive :all{-}material{-}questions{-}covered}
\NormalTok{  :from (session{-}state :answered :deferred))}

\NormalTok{(infer :next{-}unresolved{-}sub{-}question}
\NormalTok{  :from (question deep{-}dives dialogues session{-}state/answered))}
\end{Highlighting}
\end{Shaded}

\texttt{derive} denotes computation over already available workflow
state. Its semantics are deterministic, testable, and replayable.
\texttt{infer} denotes mediated LLM judgment. It requires declared
context, a prompt or intent, an expected return type, validation,
persistence of the result, and an explicit capability policy for any
external action or tool use.

Non-LLM nondeterminism, such as randomized algorithms or stochastic
simulations, is not \texttt{infer}; it can be treated as \texttt{derive}
when the seed, inputs, algorithm version, and replay policy are
explicit. When those conditions cannot be made explicit, a future
semantic refinement may need a separate primitive.

An \texttt{infer} site may return a proposed DSL expression, but that
does not give the LLM authority over the workflow. Work on LLM-assisted
DSL generation consistently points back to surrounding machinery:
grammars, parsing checks, validation loops, and execution policy
(\citeproc{ref-dslxpert2024}{\emph{{DSL-Xpert}} 2024};
\citeproc{ref-dslxpert2025}{{``{DSL-Xpert} 2.0''} 2025};
\citeproc{ref-dslGenerationFinetuningRag2024}{\emph{A Comparative Study
of {DSL} Code Generation} 2024};
\citeproc{ref-cabot2026llmsDslDevelopment}{Cabot 2026}). In the proposed
model, the LLM output is recorded as an inference result or candidate
object; deterministic validation, policy, and review decide whether it
becomes executable workflow structure.

A derivation may depend on an inference, but the inference should remain
visible as a dependency rather than being absorbed into the derived
value. For example, a workflow might infer candidate claims from a
document and then derive normalized claim objects from the inference
record. The derived object remains inspectable because its
judgment-mediated premise is explicit.

The \texttt{derive} / \texttt{infer} split is a semantic and
architectural boundary, not a security proof. The LLM may fill values
for declared inference sites, and those values may influence declared
later branches. The executor retains direct authority to apply
transition rules and trigger permitted actions by walking the workflow
object graph. When an inference result influences a later branch, the
workflow instance records a relation from the inference record to the
executor-applied branch. Capability gating remains an executor policy
requirement: model output is mediated before external actions are
executed (\citeproc{ref-yao2022react}{{Yao et al.} 2022};
\citeproc{ref-openai2026functionCalling}{OpenAI 2026}).

The choice between deterministic computation and mediated judgment is
not always obvious. For example, ``find the next unresolved
sub-question'' may be deterministic if open questions are explicitly
represented, but inferential if it requires judgment about what remains
materially unresolved. The DSL should force authors to make that
distinction visible.

\begin{figure}
\centering
\pandocbounded{\includegraphics[width=0.82\linewidth,height=0.7\textheight,keepaspectratio,alt={Figure 3. derive computes over available state; infer requests mediated LLM judgment whose recorded value may influence an executor-applied declared branch.}]{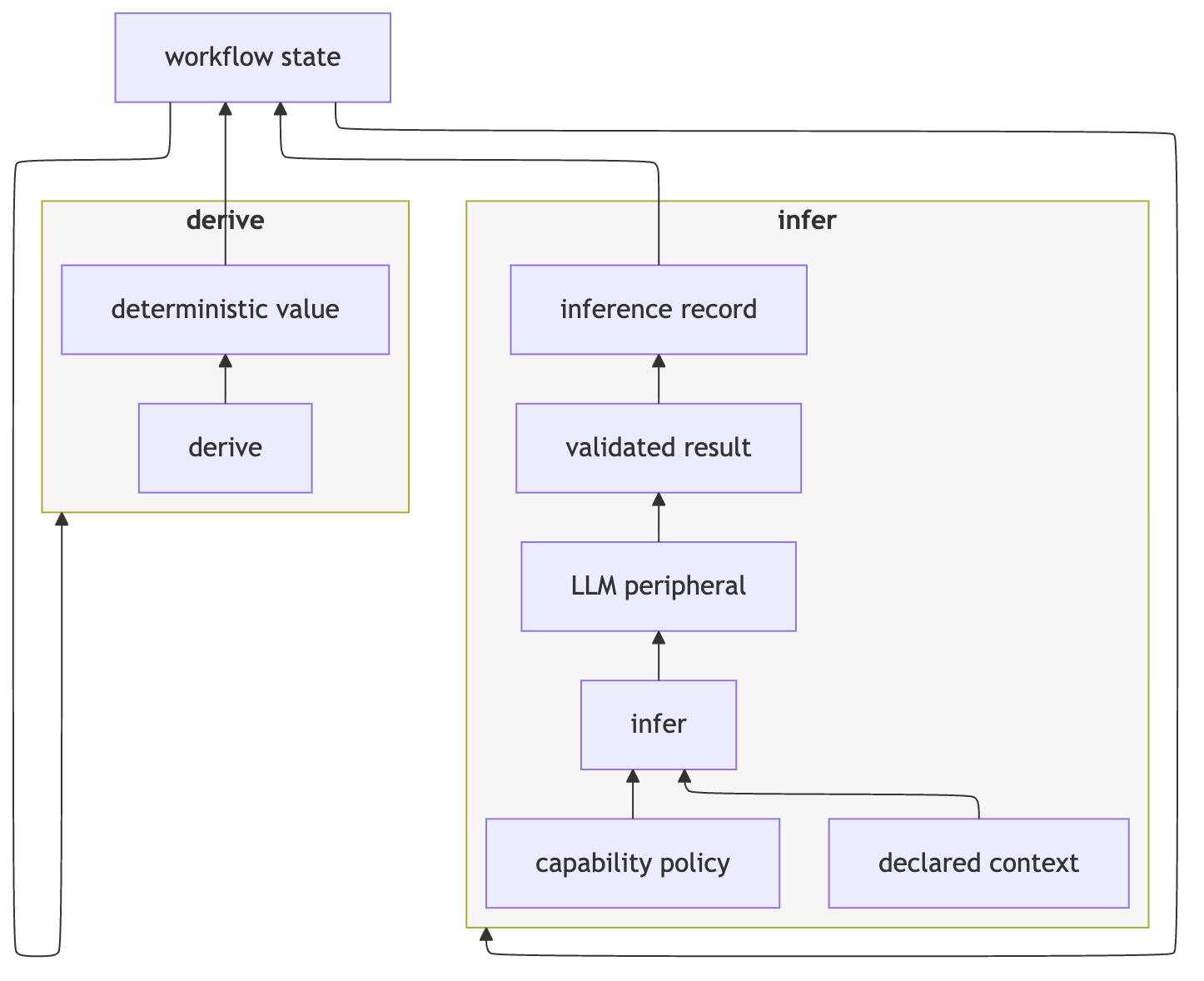}}
\caption{\protect\texttt{derive} computes over available
state; \protect\texttt{infer} requests mediated LLM judgment whose
recorded value may influence an executor-applied declared
branch.}\label{fig:mermaid-003}
\end{figure}

Every \texttt{infer} call has inspectable context. A context scope
specifies which workflow state, static resources, and executor-retrieved
material may be shown to the model service for a particular
\texttt{infer} call. The corresponding context snapshot records that
bounded view for later review.

The emphasis on explicit context is consistent with practitioner
accounts that treat context management as a first-class engineering
concern, especially when rationale and decisions must survive across
sessions rather than remain only in a conversation window
(\citeproc{ref-oreilly2026contextManagement}{O'Reilly 2026b}).

\subsection{3.5 Execution Semantics And Semantic
Checkpoints}\label{execution-semantics-and-semantic-checkpoints}

The executor is the controller within the \texttt{DSL\ machine}. In this
paper, the \texttt{DSL\ machine} names the conceptual machinery around
the language: its grammar, object constructors, validator, policy layer,
executor, and knowledge-substrate interface. The phrase deliberately
echoes Lisp-machine thinking: the point is an integrated
language-and-object environment in which forms, state, records, and
tools are interpreted together.

The executor's job is to interpret a workflow definition into a workflow
instance: it checks guards and capability policies, evaluates
\texttt{derive}, mediates \texttt{infer}, applies declared transitions,
and persists meaningful records through the knowledge-substrate
interface. This mediated structure is consistent with tool-calling APIs
in which an application supplies tools, receives model-requested calls,
performs application-side execution, and returns structured results
(\citeproc{ref-openai2026functionCalling}{OpenAI 2026}).

\begin{figure}
\centering
\pandocbounded{\includegraphics[width=0.82\linewidth,height=0.7\textheight,keepaspectratio,alt={Figure 4. The executor as controller of workflow instantiation, policy checks, runtime-mediated resources, transitions, and knowledge-substrate persistence.}]{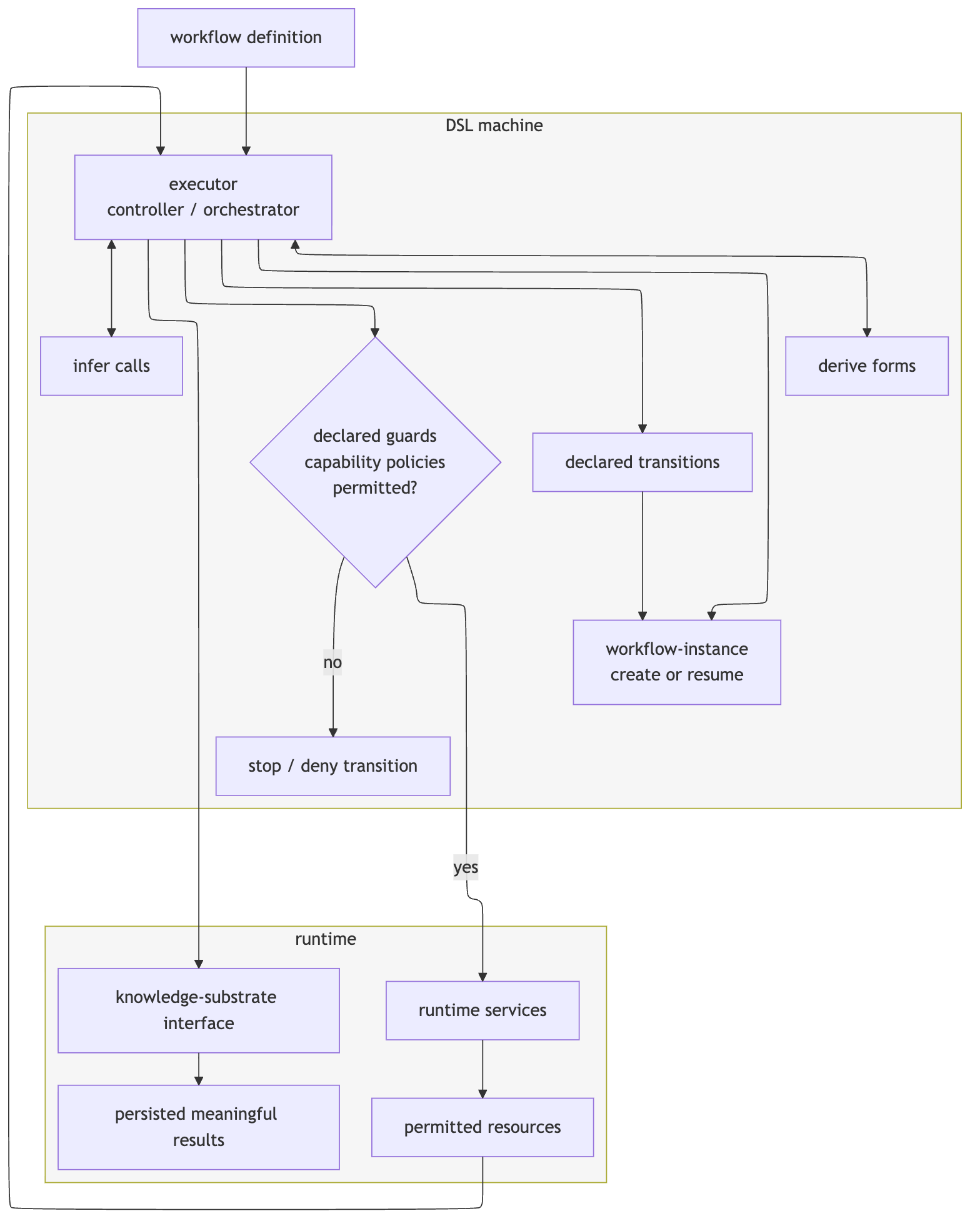}}
\caption{The executor as controller of workflow instantiation,
policy checks, runtime-mediated resources, transitions, and
knowledge-substrate persistence.}\label{fig:mermaid-004}
\end{figure}

When a workflow runs, the executor creates or resumes a
\texttt{workflow-instance} object. The instance stores live session
state, points to its workflow definition, and links to the knowledge
objects it has read or produced. If execution is interrupted, the
instance remains in the knowledge substrate and may serve as a semantic
checkpoint.

An input may be pushed by a user or retrieved by the executor under
declared context and capability rules; it is not obtained directly by
the model. Correspondingly, a model response is returned to the executor
as a candidate value or action proposal. Only after type validation and
policy checks may it be persisted as an inference record, used in an
executor-applied declared branch, or result in a permitted external
effect.

Runtime services may involve communication with external model or tool
processes. In the current model, any such communication that changes
workflow state, context, records, or permitted effects is mediated by
the executor. Explicit message, event, or channel objects are not part
of the present core vocabulary.

Workflow instances as persistent objects can support narrow resumption
rules. When the executor reaches a step whose prior result may still
apply, it may query the knowledge substrate for a compatible record
associated with the current workflow instance, step, and subject. A
prior \texttt{panel-record} is one example: if it exists with a recorded
choice and satisfies the workflow's compatibility policy, the executor
may reuse that choice instead of presenting the same deliberation again.
This is a scoped reuse rule, not a general claim that repeated
interactions are interchangeable.

The DSL should avoid general \texttt{goto}-like flow. For the target
class of workflows, local structural flow is sufficient: continue,
repeat, stop, accept, reject, defer. More complex routing should be
handled by composing smaller workflows rather than by adding arbitrary
jumps inside a workflow.

Avoiding arbitrary jumps keeps workflows inspectable as data and reduces
the risk that the DSL becomes a second general-purpose programming
language hidden inside the object model.

For a more concrete account of the proposed model, Appendix A sketches
primitive read/write behavior over workflow state and the knowledge
substrate, while Appendix B gives illustrative notation and a worked
workflow; both are explanatory sketches, not a formal language
specification or implementation.

\subsection{3.6 Contrast With Execution-Oriented
Approaches}\label{contrast-with-execution-oriented-approaches}

The distinction between this proposal and current workflow systems is
not primarily syntactic. It is ontological: what kind of thing is a
workflow, and what happens to it after it runs?

AgentSPEX represents a strong execution-oriented approach. It makes
control flow explicit, supports typed steps, manages state, and provides
checkpointing, verification, and logging. This addresses the execution
control problem: how to make agent behavior more predictable,
inspectable, and maintainable during a run.

This paper instead addresses the knowledge persistence problem: how to
represent model-mediated judgments, human decisions, derived results,
and their contexts so that they remain available as knowledge after
execution. Provenance systems show that execution-related artifacts can
be modeled for later inspection, but the proposed model gives selected
workflow artifacts semantic roles before any provenance export. In this
model, semantic persistence means retaining typed workflow objects and
inference records, their relations, and their reviewable contexts inside
the same substrate.

The difference is not that a workflow engine, database, or provenance
system cannot store related artifacts. It is that the model assigns
declared semantic roles to workflow definitions, workflow instances,
inference records, and related objects before execution begins treating
them as stored data. The proposal is therefore not a replacement for
provenance vocabularies; it is a semantic object model whose records
could later be mapped to provenance representations.

In that sense, the model is provenance-compatible but not
provenance-complete. Several proposed objects compare naturally with
PROV entities, activities, plans, collections, and relations. The model
adds what PROV-DM does not define: DSL-level roles, executor mediation,
infer boundaries, approval authority, panel structure, and lifecycle
questions.

{\def\LTcaptype{none} % do not increment counter
\begin{longtable}[]{@{}
  >{\raggedright\arraybackslash}p{(\linewidth - 6\tabcolsep) * \real{0.2153}}
  >{\raggedright\arraybackslash}p{(\linewidth - 6\tabcolsep) * \real{0.2349}}
  >{\raggedright\arraybackslash}p{(\linewidth - 6\tabcolsep) * \real{0.2349}}
  >{\raggedright\arraybackslash}p{(\linewidth - 6\tabcolsep) * \real{0.2349}}@{}}
\toprule\noalign{}
\begin{minipage}[b]{\linewidth}\raggedright
Approach family
\end{minipage} & \begin{minipage}[b]{\linewidth}\raggedright
Primary unit
\end{minipage} & \begin{minipage}[b]{\linewidth}\raggedright
What persistence usually captures
\end{minipage} & \begin{minipage}[b]{\linewidth}\raggedright
What this paper adds
\end{minipage} \\
\midrule\noalign{}
\endhead
\bottomrule\noalign{}
\endlastfoot
Provenance models & Entities, activities, agents, and influence
relations & Histories of data, activities, decisions, and derivations,
typically captured after the fact & Declares \texttt{infer}, approval,
and panel as DSL-level roles before any provenance export, rather than
relying only on a later provenance encoding \\
Notebook and hypertext history & Cells, outputs, links, anchors, and
navigable histories & Exploratory execution traces and linked
information structures & Persists the workflow definition and running
instance themselves as typed objects, not only their outputs and
links \\
Agent workflow/\allowbreak{}checkpoint systems & Graphs, steps, state, checkpoints,
modules, and tool calls & Execution persistence: state, recovery points,
traces, logs, and outputs & Persists mediated roles, specifically
inference, approval, and panel records, as queryable semantic objects
that outlive the run, not only resumable execution state \\
Agent harness practice & Context assembly, tools, permissions, feedback,
observability, and recovery & Operational control around model use,
often implemented in surrounding code & Makes those operational roles
declared, typed language constructs rather than implicit harness
behavior \\
Proposed model & Workflow objects and mediated records in a knowledge
substrate & Semantic persistence: typed objects and relations retained
beyond execution & Combines explicit \texttt{derive} /\allowbreak{} \texttt{infer}
mediation with object persistence \\
\end{longtable}
}

\textbf{Table 3.} Approach families compared by their primary unit of
persistence and what each typically retains. The comparison is at the
level of approach categories, not individual systems; Appendix C gives a
more granular object-level mapping against PROV-DM, and lifecycle and
supersession semantics remain Future Work.

This distinction generalizes beyond any single framework. The dominant
pattern is workflow as configuration, execution as process, results as
output. The proposed pattern is workflow as object, execution as
instantiation, results as knowledge citizens.

\section{4. Exploratory Vocabulary
Scan}\label{exploratory-vocabulary-scan}

The preceding sections propose a core semantic schema and vocabulary.
This section treats a selected pilot scan of 77 skill-like workflow
artifacts as a vocabulary-design probe. The corpus was selected,
qualitative, and heterogeneous, combining public agent and prompt
workflow artifacts, operational skill artifacts, and local/private
design material. Appendix D gives the corpus grouping, primitive-count
table, and source-use cautions. The counts in Appendix D are descriptive
of that selected corpus only; they are used here to identify vocabulary
pressure points rather than to estimate prevalence. The scan should
therefore be read as design feedback for the vocabulary, not as
empirical validation of the thesis.

Within that limit, many scanned artifacts contained structured
resources, guards, context, steps, branching, deterministic operations,
judgment-based operations, and records.

The value of the scan is vocabulary refinement. In particular, synthesis
of the scanned artifacts suggested separating \texttt{approval} from
deliberative \texttt{panel}, although \texttt{approval} was not part of
the original scoring checklist and therefore has no corpus count in
Appendix D. The scan also suggests distinguishing different forms of
\texttt{context} (source-authority, inference-visible, operating, and
progressive-disclosure), and separating active \texttt{state} from
durable \texttt{record}. It also suggests possible future primitives or
annotations for \texttt{capability} or \texttt{action},
\texttt{handoff}, \texttt{promotion}, and compatible resumption
behavior.

\section{5. Discussion And Limits}\label{discussion-and-limits}

The proposed model offers three affordances.

First, it makes workflow history queryable. A user can ask for all
deferred panels, all decisions based on a particular document, or all
workflows that touched a given assumption.

Second, it makes later review more precise. Because every \texttt{infer}
has explicit context and every panel records a context snapshot, later
review can inspect the information available when a decision was made.

Third, it clarifies authority and influence. Deterministic computation,
LLM judgment, user approval, and executor-applied control flow are
separate concepts in the language; the proposed records make those
authority and influence relations inspectable.

The claim is not that more retention is better, but that consequential
retained material should have explicit type, context, identity, and
dependency relations. Whether these affordances improve audit quality or
trust in practice is an empirical question, not a claim of this paper.

The model also has costs. Retained inference, approval, panel, and
context records consume storage and increase index, lifecycle, and
review complexity. More importantly, not every intermediate artifact
deserves durable prominence: model output may be redundant, low-quality,
uncertain, later disputed, or actively misleading.

The conceptual model does not yet choose a record-lifecycle policy. An
implementation would need explicit rules for retaining, compressing,
superseding, or deleting records, and would need to state what
provenance survives each operation. Authorship, credit assignment, panel
interpretation, and contestability are therefore governance and
evaluation questions, not storage-format details.

The \texttt{derive}/\texttt{infer} distinction also requires discipline.
Some tasks can be implemented either way. The DSL should not pretend the
boundary is mathematically obvious in all cases. Instead, it should make
the author's choice explicit and inspectable. When a result changes
because a judgment has been reviewed, the model can preserve the old
inference record, create a new one, link the two by a supersession or
dispute relation, and re-derive any dependent objects with the changed
premise visible.

The evaluation question is how well these records support review tasks.
Future work should test inspection, attribution, audit, and
reproducibility claims rather than infer them from persistence alone.

\section{6. Future Work}\label{future-work}

Future work falls into three main categories.

First, the semantic model needs refinement. The next version should
specify the operational semantics of \texttt{derive}, \texttt{infer},
guards, approvals, context snapshots and resumption. A later version
should also decide whether non-LLM nondeterministic operations, such as
randomized algorithms, stochastic simulations, or external measurements,
are adequately covered by replay-controlled \texttt{derive} or require a
separate semantic primitive. The limited semantic sketch in Appendix A
should be extended into formal transition rules before the model is used
to claim semantic-persistence guarantees. It should also define record
lifecycle policy: when records are retained, summarized, compacted,
archived, deleted, or tombstoned, and what provenance survives each
operation. A parallel governance track should define threat models,
capability boundaries, and review authority.

Second, the model needs evidence and comparison. A larger scan could
test whether the proposed primitive vocabulary travels beyond selected
skill-like artifacts into runbooks, issue workflows, research protocols,
and operational procedures. Future work should compare semantic workflow
persistence with checkpoint-based workflow systems, scientific
provenance models, notebook provenance, and agent-provenance models. A
formal W3C PROV export or a template-based provenance representation is
one concrete comparison path (\citeproc{ref-w3c2013prov}{W3C 2013};
\citeproc{ref-singh2019decisionProvenance}{Singh et al. 2019}). Separate
evaluation should examine governance, trust effects, authorship
attribution, reproducibility, performance, and local-first inference.

Third, possible implementations should remain small and explicit. One
possible Common Lisp path would begin with \texttt{define-workflow} and
related forms as macros that construct CLOS object graphs rather than
opaque executable code (\citeproc{ref-bobrow1988closSpec}{Bobrow et al.
1988}; \citeproc{ref-gabriel1991clos}{Gabriel et al. 1991}). A minimal
Common Lisp read-eval-print loop (\texttt{REPL}) prototype could then
define a workflow, start an instance, evaluate \texttt{derive}, dispatch
\texttt{infer} through a pluggable model adapter, and show the resulting
state and records. Durable persistence, query operations, optional
serialization, and cacheable \texttt{infer} records could be added after
the object semantics are stable. A prior inference may be reusable when
its context, version, and freshness satisfy an explicit compatibility
policy, but reuse should not be treated as proof of truth
(\citeproc{ref-ding2025vcache}{{Ding et al.} 2025}). More reflective
implementation paths could later draw on metaobject protocol techniques,
but they are not required for the conceptual architecture
(\citeproc{ref-kiczales1991amop}{Kiczales et al. 1991}).

After these three categories, architecture-adjacent work could explore
externalized context environments, recursive decomposition strategies,
and heterogeneous multi-agent systems
(\citeproc{ref-recursiveLanguageModels2025}{\emph{Recursive Language
Models} 2025}; \citeproc{ref-recursivemas2026}{\emph{Recursive
Multi-Agent Systems} 2026}). Because definitions and instances are
first-class objects, it could also examine meta-workflows that
instantiate, coordinate, review, or promote results from other workflows
through explicit \texttt{handoff}, \texttt{promotion}, and capability
relations. Explicit message, event, or channel objects for distributed
coordination would belong to that refinement track.

A first evaluation should remain deliberately small. A prototype could
implement one workflow definition, one resumable instance, typed
inference/approval/panel records, and explicit context snapshots. It
would compare its recorded history with a checkpoint-and-trace baseline,
measure record volume and indexing cost, and test alternative retention
policies. Separate human evaluation would be required for questions of
attribution, provenance usefulness, review quality, or reproducibility.

\section{7. Conclusion}\label{conclusion}

LLM workflow systems have made important progress by making control flow
explicit. This paper argues for a complementary Lisp-inspired model in
which workflows and their consequential inference, context, and
dependency records inhabit a shared knowledge substrate rather than
remain only external traces of execution.

The model has four central consequences. First, workflow definitions are
represented as semantic data objects rather than only executable
procedures or external configuration. Second, workflow instances persist
as resumable occurrences with inspectable state. Third, deterministic
\texttt{derive} operations remain distinct from mediated \texttt{infer}
judgments and their recorded dependencies. Fourth, approvals and
deliberations are retained as typed records with context snapshots.

The added value of the Lisp-inspired notation is that written forms can
stay close to the object structure they denote. Workflow execution would
not merely produce knowledge and then disappear: under \emph{semantic
persistence}, the \emph{workflow itself becomes knowledge}, with its
definition, its instances, and the inference and deliberation records
that shaped them retained as inspectable objects in the same substrate
as the knowledge they produce.

\section{Acknowledgements}\label{acknowledgements}

Language-model-assisted research and writing tools were used during the
preparation of this manuscript for organizing author input, exploring
alternative formulations, structured critique, project-record
maintenance, and prose refinement. In the terms of this paper, these
uses were treated as mediated inference steps rather than sources of
authority or independent agency. The authors reviewed, revised, and
approved the final manuscript, verified the cited sources, and retain
full responsibility for the article's claims, argument, wording, and
references.

The broader research process also used Git commits and local post-commit
traces to preserve parts of the project history. These traces informed
the authors' thinking about workflow persistence and inference records,
but are not presented as empirical evidence in this article.

\section{Appendix A. Limited Semantic
Sketch}\label{appendix-a.-limited-semantic-sketch}

This appendix gives a transition-oriented sketch of the model's
operational vocabulary. It identifies the state, substrate, context,
policy, value, and record positions that a later formal treatment would
make precise, while the present paper uses them to clarify the proposed
object model.

In this sketch, workflow state means the active execution state of a
running workflow instance: current bindings, step position, temporary
values, and pending transitions. Knowledge state means the persistent
semantic objects and relations retained in the knowledge substrate:
workflow definitions, workflow instances, records, context snapshots,
dependency links, and supersession links. A primitive may affect one,
both, or neither; all such effects are mediated by the executor.

\subsection{A.1 Names Used In This
Appendix}\label{a.1-names-used-in-this-appendix}

The sketch uses a small set of names for objects already introduced in
the article.

Workflow objects:

\begin{itemize}
\tightlist
\item
  \texttt{D}: workflow-definition, the declared process object.
\item
  \texttt{I}: workflow-instance, one running or resumable occurrence of
  \texttt{D}.
\end{itemize}

State and substrate:

\begin{itemize}
\tightlist
\item
  \texttt{S}: workflow-state, the active execution state of \texttt{I}.
\item
  \texttt{S\textquotesingle{}}: successor workflow-state after an
  executor-applied step.
\item
  \texttt{K}: knowledge substrate before the step.
\item
  \texttt{K\textquotesingle{}}: successor knowledge substrate after
  persisted objects or relations are added.
\end{itemize}

Context and policy:

\begin{itemize}
\tightlist
\item
  \texttt{C}: context scope, the declared rule for material visible to
  an operation.
\item
  \texttt{CS}: context snapshot, the bounded view actually assembled for
  later review.
\item
  \texttt{P}: executor-side policy, including guards, validation,
  capability, or compatibility rules.
\end{itemize}

Values and records:

\begin{itemize}
\tightlist
\item
  \texttt{e}: deterministic expression used by \texttt{derive}.
\item
  \texttt{q}: prompt or intent used by \texttt{infer}.
\item
  \texttt{raw}: candidate model output before validation.
\item
  \texttt{v}: value accepted by executor validation.
\item
  \texttt{r}: rejection, failed validation, policy denial, or
  unavailable result.
\item
  \texttt{R}: typed persistent record or relation.
\end{itemize}

\subsection{A.2 Control Constructs And Primitive
Effects}\label{a.2-control-constructs-and-primitive-effects}

The workflow-definition contains declared control constructs such as
\texttt{step}, \texttt{branch}, and \texttt{loop}. A \texttt{step} names
a unit of execution, a \texttt{branch} selects among declared
transitions, and a \texttt{loop} represents bounded repetition, retry,
review, or batch iteration. These constructs are part of the DSL, rather
than hidden in Python glue code, framework callbacks, or an external
orchestration graph. In this sketch, \texttt{loop} is treated as a
structural control construct that organizes repeated primitive effects,
not as a separate primitive with its own effect schema; Appendix D
reports it because the scan checklist counted explicit iteration
patterns.

The primitives below describe effects that occur within those declared
constructs. In current harness and workflow systems, similar behavior
often appears as graph nodes, edges, retry loops, interrupt points,
scripts, or callback code. Here it is represented inside the
workflow-definition object interpreted by the executor. When a branch
depends on an \texttt{infer} result, the branch remains
executor-applied, and the dependency from the inference record to the
branch should remain visible.

\begin{center}
\pandocbounded{\includegraphics[width=0.82\linewidth,height=0.62\textheight,keepaspectratio]{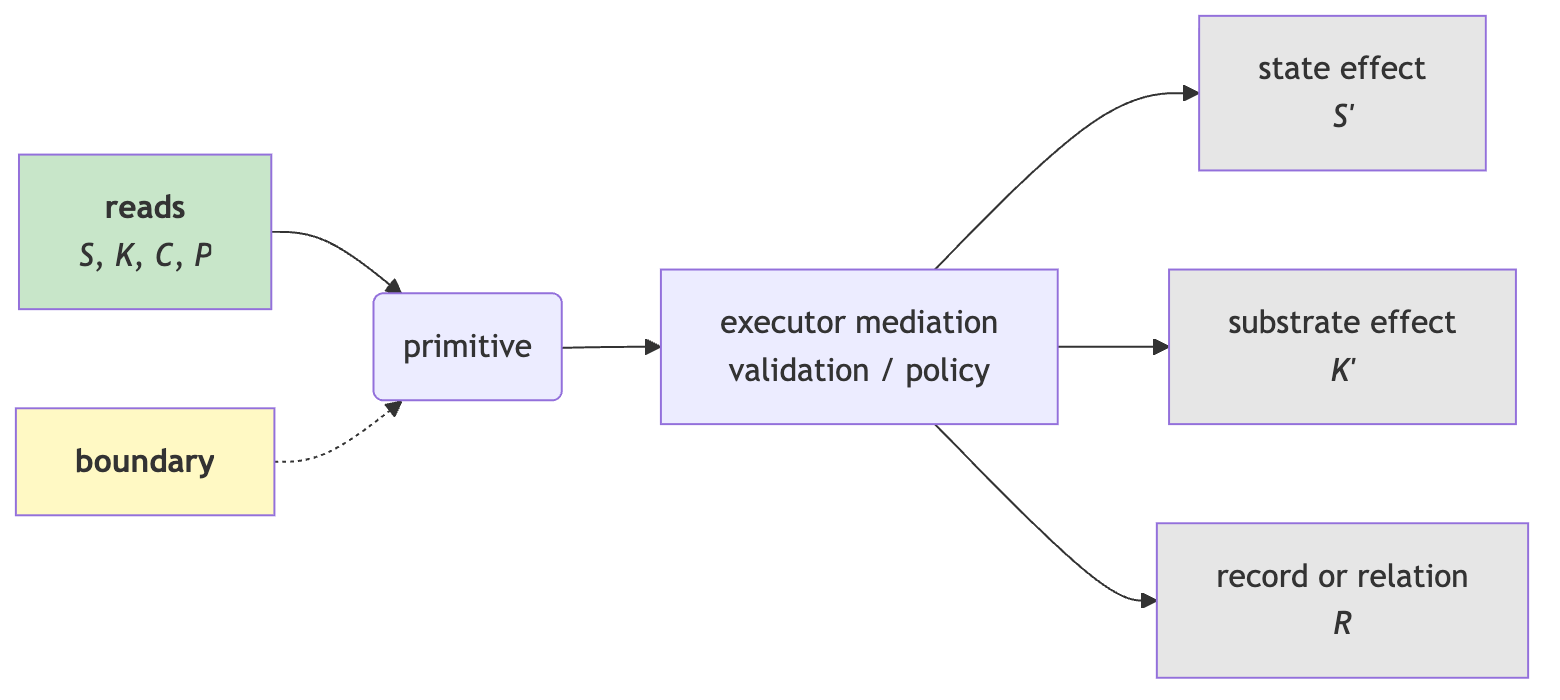}}
\end{center}

Each primitive specializes this pattern. Some primitives do not read
every named object, and some leave either active state or the knowledge
substrate unchanged.

\subsubsection{\texorpdfstring{\texttt{resource}}{resource}}\label{resource}

\begin{quote}
Makes explicit the files, repositories, APIs, tool registries, RAG
indexes, MCP connectors, memory files, or configured skill inputs that a
harness would otherwise load procedurally.
\end{quote}

Given a declared resource reference and policy \texttt{P}, the executor
resolves permitted material; it may bind that material into
\texttt{S\textquotesingle{}}; it may persist a \texttt{resource-binding}
or \texttt{retrieval-record} in \texttt{K\textquotesingle{}}.

\begin{center}
\pandocbounded{\includegraphics[width=0.82\linewidth,height=0.62\textheight,keepaspectratio]{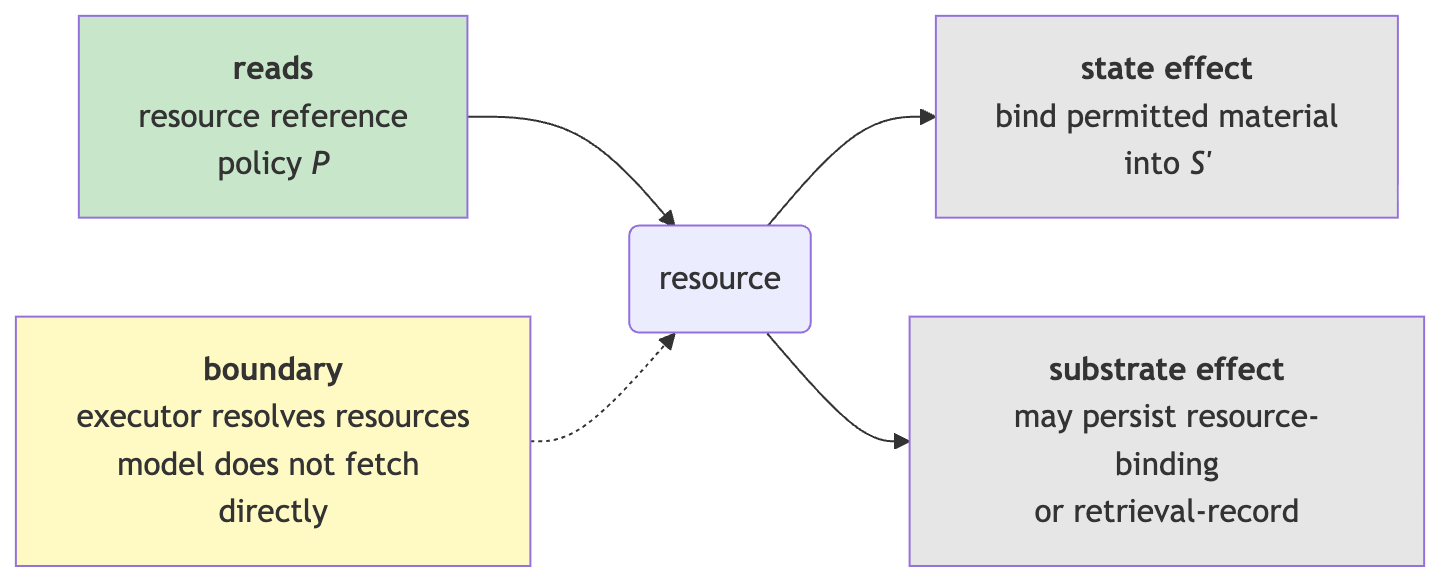}}
\end{center}

\subsubsection{\texorpdfstring{\texttt{context}}{context}}\label{context}

\begin{quote}
Captures what current systems handle through prompt assembly,
context-window management, AGENTS-style instructions, sub-agent context
isolation, progressive disclosure, or retrieved snippets.
\end{quote}

Given workflow-state \texttt{S}, knowledge substrate \texttt{K}, and
context scope \texttt{C}, the executor assembles bounded visible
material; it may persist context snapshot \texttt{CS} in
\texttt{K\textquotesingle{}}.

\begin{center}
\pandocbounded{\includegraphics[width=0.82\linewidth,height=0.62\textheight,keepaspectratio]{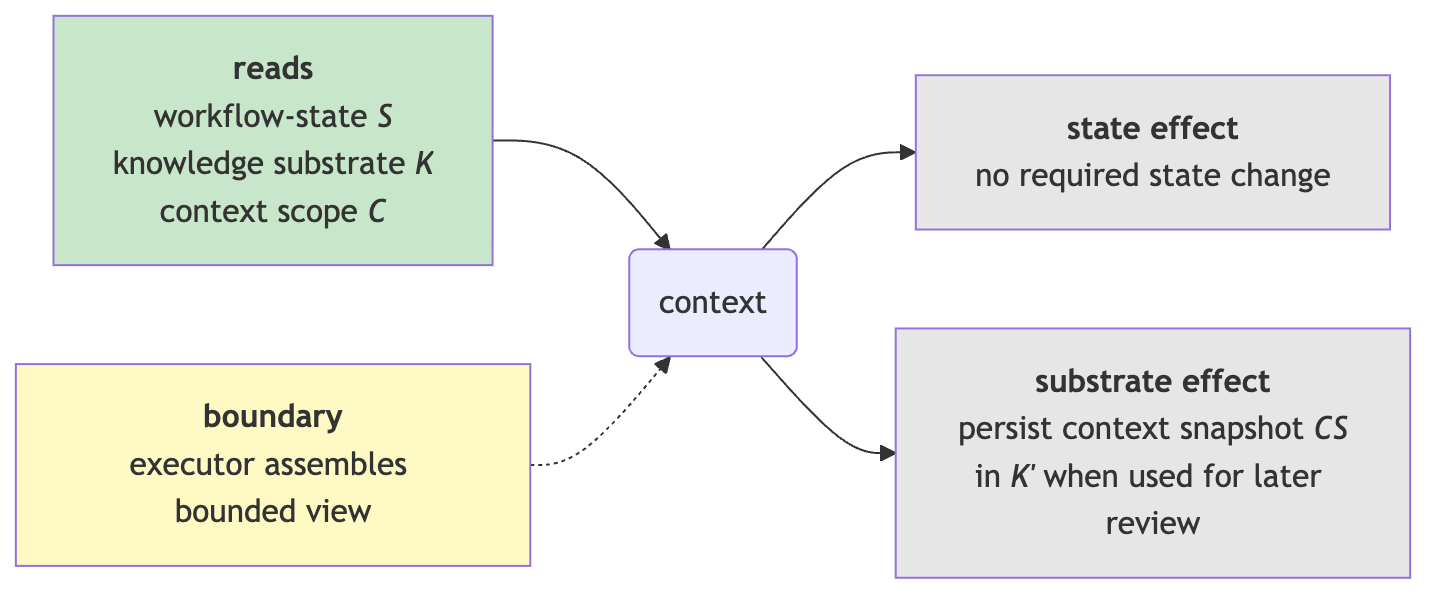}}
\end{center}

\subsubsection{\texorpdfstring{\texttt{guard}}{guard}}\label{guard}

\begin{quote}
Represents scope locks, permissions, sandbox rules, preflight checks,
hook conditions, and stop rules that harnesses often enforce outside the
model.
\end{quote}

Given workflow-state \texttt{S} and policy condition \texttt{P}, the
executor evaluates whether a transition is permitted; it may continue,
stop, or deny the transition; it may persist a guard result or violation
record in \texttt{K\textquotesingle{}}.

\begin{center}
\pandocbounded{\includegraphics[width=0.82\linewidth,height=0.62\textheight,keepaspectratio]{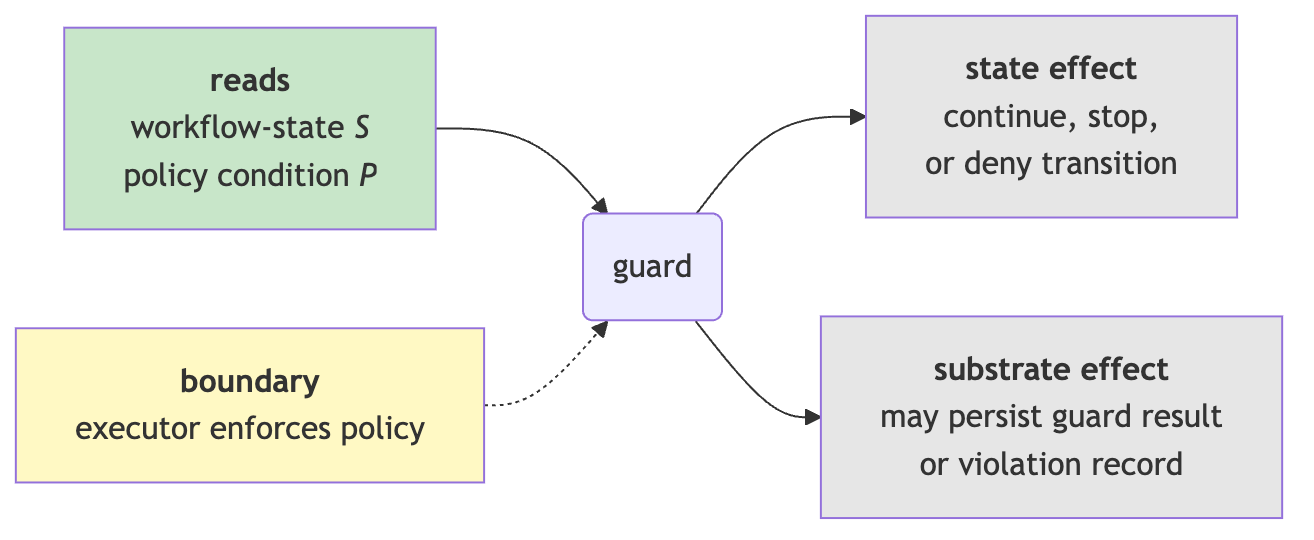}}
\end{center}

\subsubsection{\texorpdfstring{\texttt{derive}}{derive}}\label{derive}

\begin{quote}
Represents deterministic harness work: scripts, parsers, schema checks,
linters, tests, type checks, formatters, routing rules, or other
computational feedback controls.
\end{quote}

Given workflow-state \texttt{S} and deterministic expression \texttt{e},
evaluate \texttt{e} over \texttt{S} to produce value \texttt{v}; the
executor may bind \texttt{v} into \texttt{S\textquotesingle{}}; it may
persist \texttt{derived-object(v)} and dependency links in
\texttt{K\textquotesingle{}}.

\begin{center}
\pandocbounded{\includegraphics[width=0.82\linewidth,height=0.62\textheight,keepaspectratio]{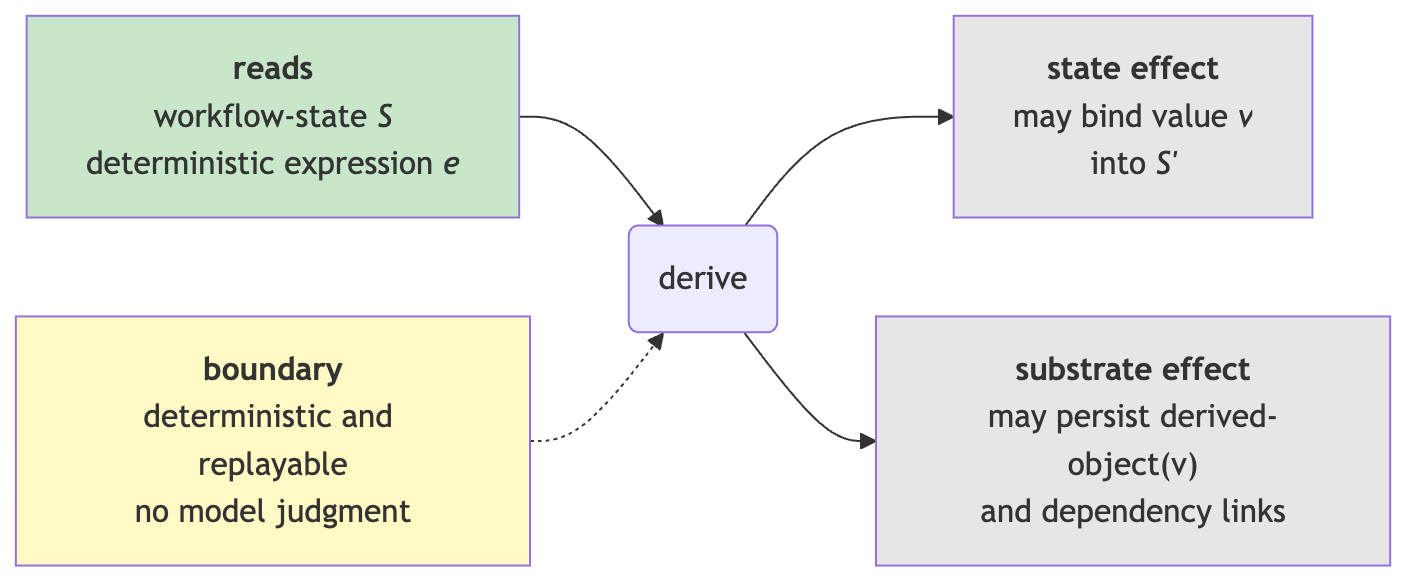}}
\end{center}

\subsubsection{\texorpdfstring{\texttt{infer}}{infer}}\label{infer}

\begin{quote}
Represents model-mediated judgment: classification, synthesis, drafting,
semantic review, LLM-as-judge, or candidate DSL generation, with
executor validation before effects.
\end{quote}

Given workflow-state \texttt{S}, context scope \texttt{C}, prompt or
intent \texttt{q}, and policy \texttt{P}, the executor assembles context
snapshot \texttt{CS}; the model service returns candidate output
\texttt{raw}; the executor validates \texttt{raw} to produce value
\texttt{v} or rejection \texttt{r}; it may persist an
\texttt{inference-record} in \texttt{K\textquotesingle{}}; it may bind
\texttt{v} into \texttt{S\textquotesingle{}} only through a declared
transition.

\begin{center}
\pandocbounded{\includegraphics[width=0.82\linewidth,height=0.62\textheight,keepaspectratio]{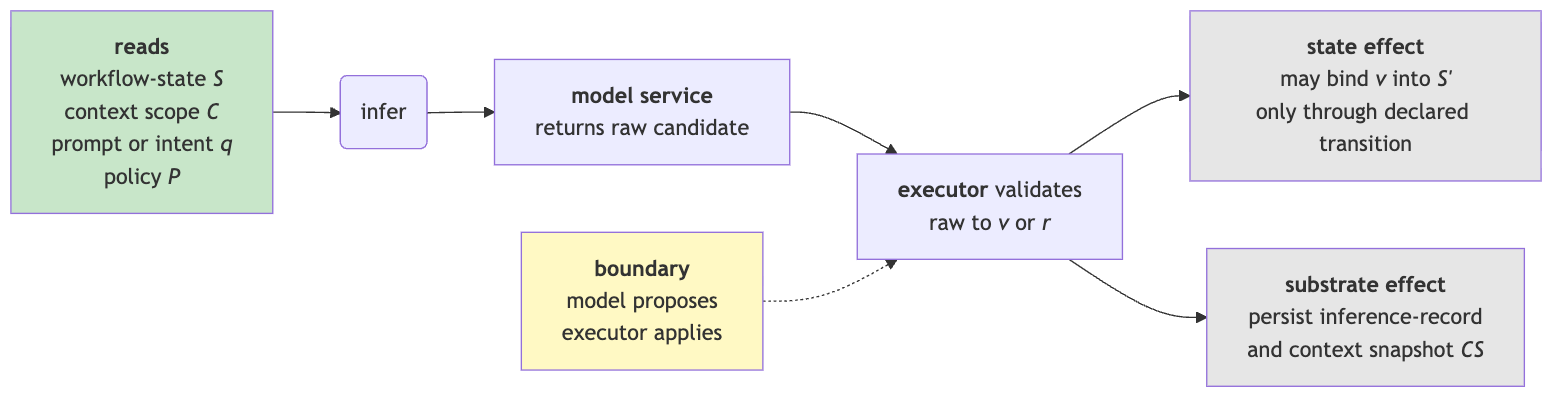}}
\end{center}

\subsubsection{\texorpdfstring{\texttt{approval}}{approval}}\label{approval}

\begin{quote}
Represents HITL approval gates, human confirmation prompts,
interrupt/resume checkpoints, or policy authorization before a
transition or side effect.
\end{quote}

Given a request, context snapshot \texttt{CS}, and authority policy
\texttt{P}, user or policy authority returns approve, reject, or defer;
the executor may gate the next transition; it persists an
\texttt{approval-record} in \texttt{K\textquotesingle{}}.

\begin{center}
\pandocbounded{\includegraphics[width=0.82\linewidth,height=0.62\textheight,keepaspectratio]{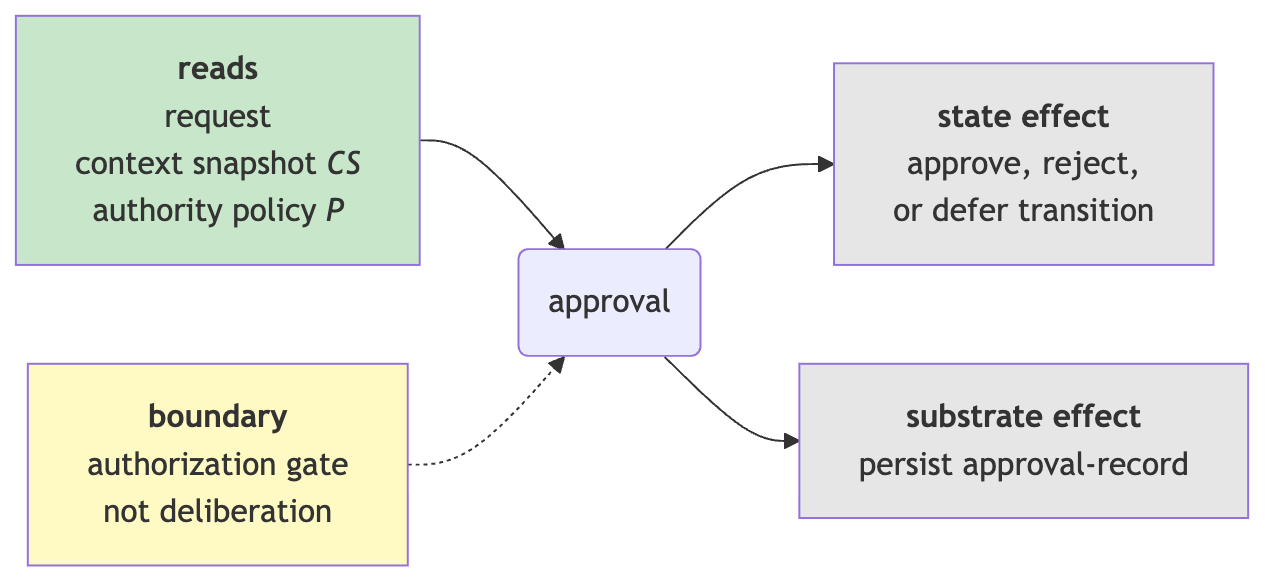}}
\end{center}

\subsubsection{\texorpdfstring{\texttt{panel}}{panel}}\label{panel}

\begin{quote}
Represents structured review rather than a simple approval: author
review, expert-panel critique, options comparison, design review, or
deliberative decision capture.
\end{quote}

Given a motion, options, arguments, and context snapshot \texttt{CS},
structured deliberation produces a choice or deferral; the executor may
use that result in a later declared transition; it persists a
\texttt{panel-record} in \texttt{K\textquotesingle{}}.

\begin{center}
\pandocbounded{\includegraphics[width=0.82\linewidth,height=0.62\textheight,keepaspectratio]{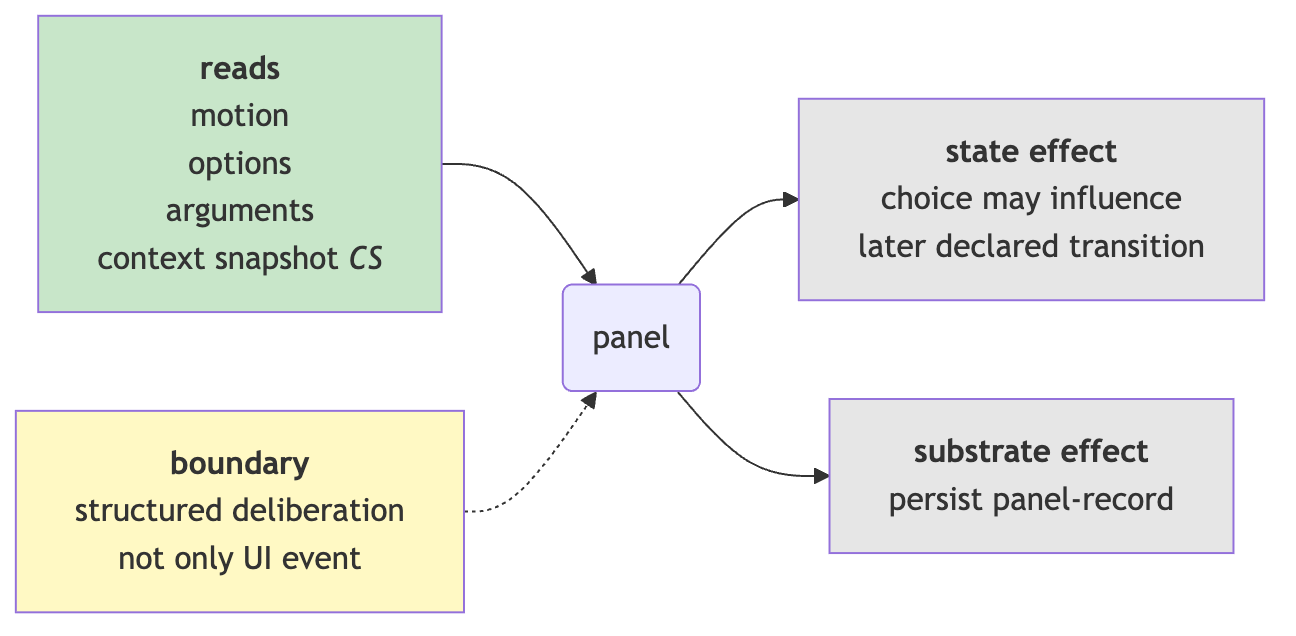}}
\end{center}

\subsubsection{\texorpdfstring{\texttt{record}}{record}}\label{record}

\begin{quote}
Turns logs, traces, reports, review notes, decisions, checkpoints,
transcripts, generated artifacts, or memory entries into typed
persistent semantic objects.
\end{quote}

Given consequential output, decision material, or context, the executor
creates a typed durable object; \texttt{S} may advance, but the durable
effect is in \texttt{K\textquotesingle{}}.

\begin{center}
\pandocbounded{\includegraphics[width=0.82\linewidth,height=0.62\textheight,keepaspectratio]{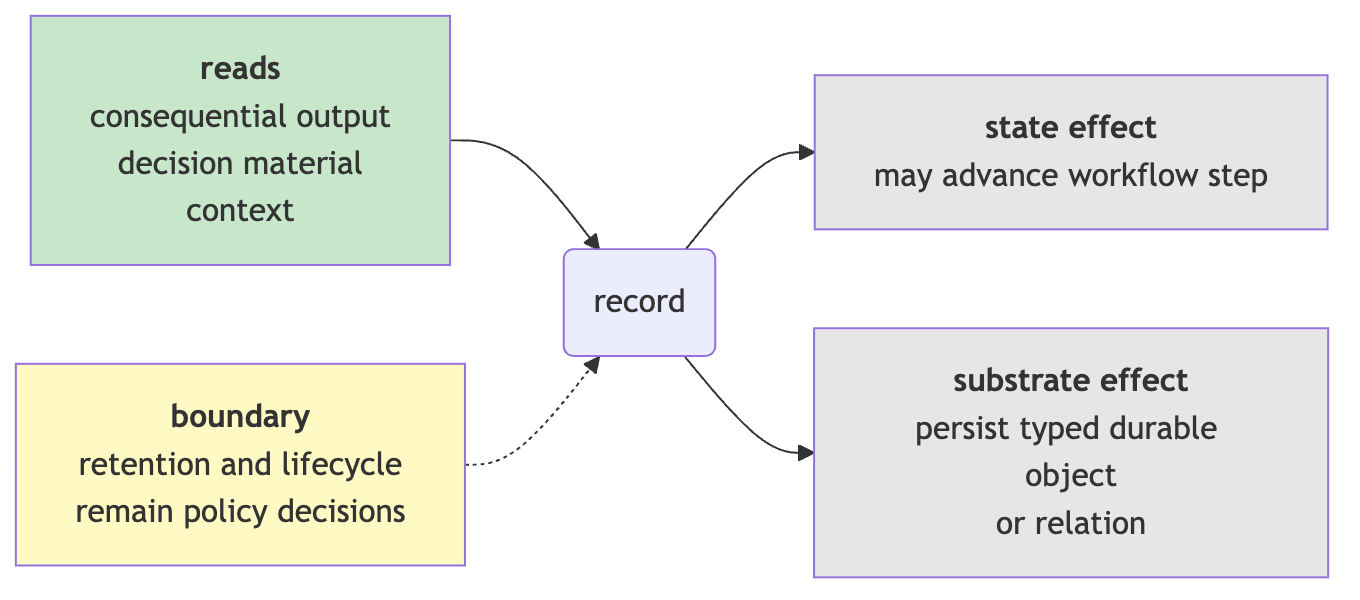}}
\end{center}

\subsubsection{\texorpdfstring{\texttt{state}}{state}}\label{state}

\begin{quote}
Names active workflow memory: graph state, checkpoint state, task-file
state, workspace state, session memory, or recovery state used while
execution continues.
\end{quote}

Given workflow-instance \texttt{I} and current bindings, the executor
reads or updates active execution memory; state becomes durable evidence
only when represented by a record in \texttt{K\textquotesingle{}}.

\begin{center}
\pandocbounded{\includegraphics[width=0.82\linewidth,height=0.62\textheight,keepaspectratio]{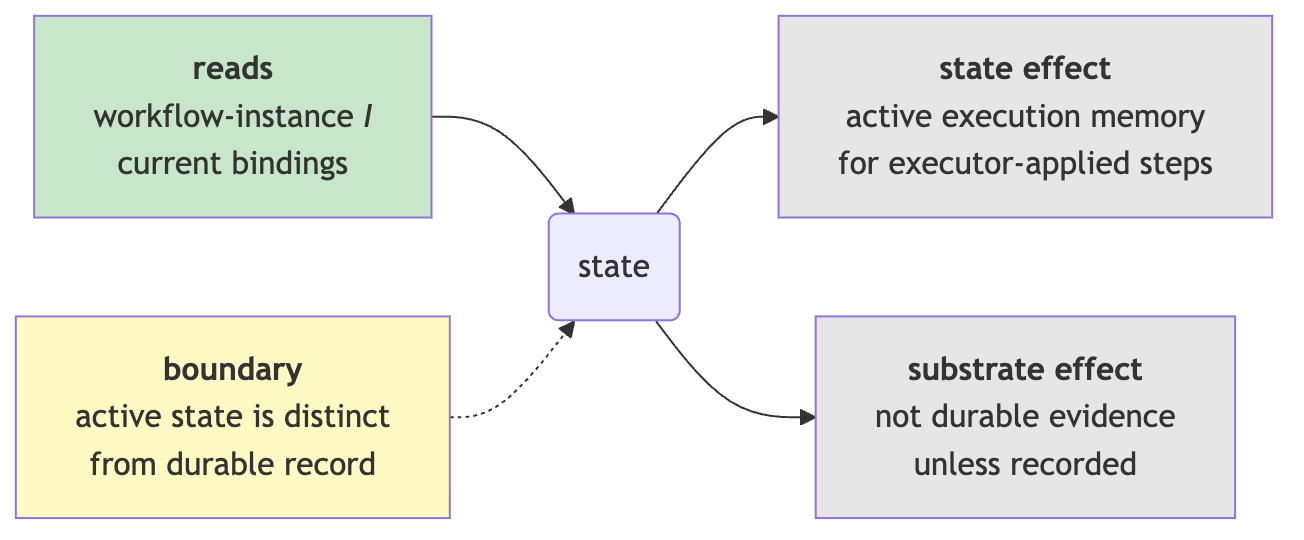}}
\end{center}

\subsubsection{\texorpdfstring{\texttt{capability} /
\texttt{action}}{capability / action}}\label{capability-action}

\begin{quote}
Candidate refinement for side-effecting tools: function calls, MCP
tools, shell commands, browser actions, API writes, deployments, or
external process calls mediated by permissions.
\end{quote}

Given a validated request and capability policy \texttt{P}, the executor
may perform or deny a side-effecting action; it may persist an action or
tool-result record in \texttt{K\textquotesingle{}}.

\begin{center}
\pandocbounded{\includegraphics[width=0.82\linewidth,height=0.62\textheight,keepaspectratio]{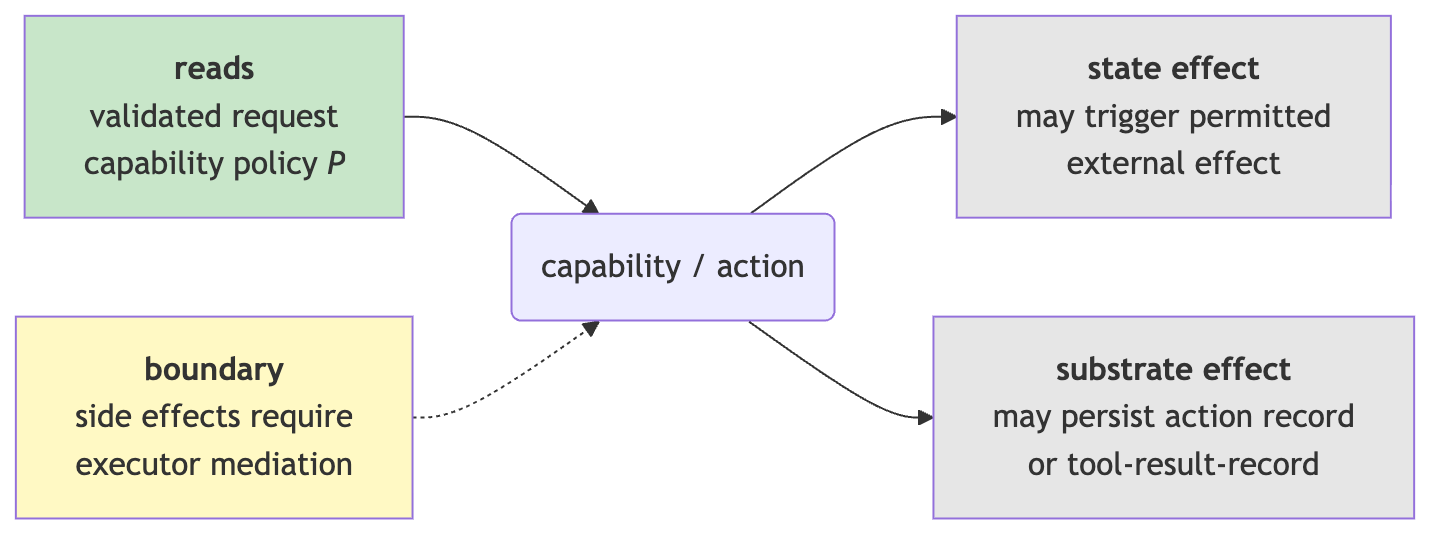}}
\end{center}

\subsubsection{\texorpdfstring{\texttt{handoff} /
\texttt{promotion}}{handoff / promotion}}\label{handoff-promotion}

\begin{quote}
Candidate refinement for delegation and elevation: sub-agent routing,
queueing work for another workflow, promoting a result to shared memory,
opening an issue, or sending material to review.
\end{quote}

Given a source object, target workflow or context, and policy
\texttt{P}, the executor may route or promote an object; it may persist
a handoff or promotion relation in \texttt{K\textquotesingle{}}.

\begin{center}
\pandocbounded{\includegraphics[width=0.82\linewidth,height=0.62\textheight,keepaspectratio]{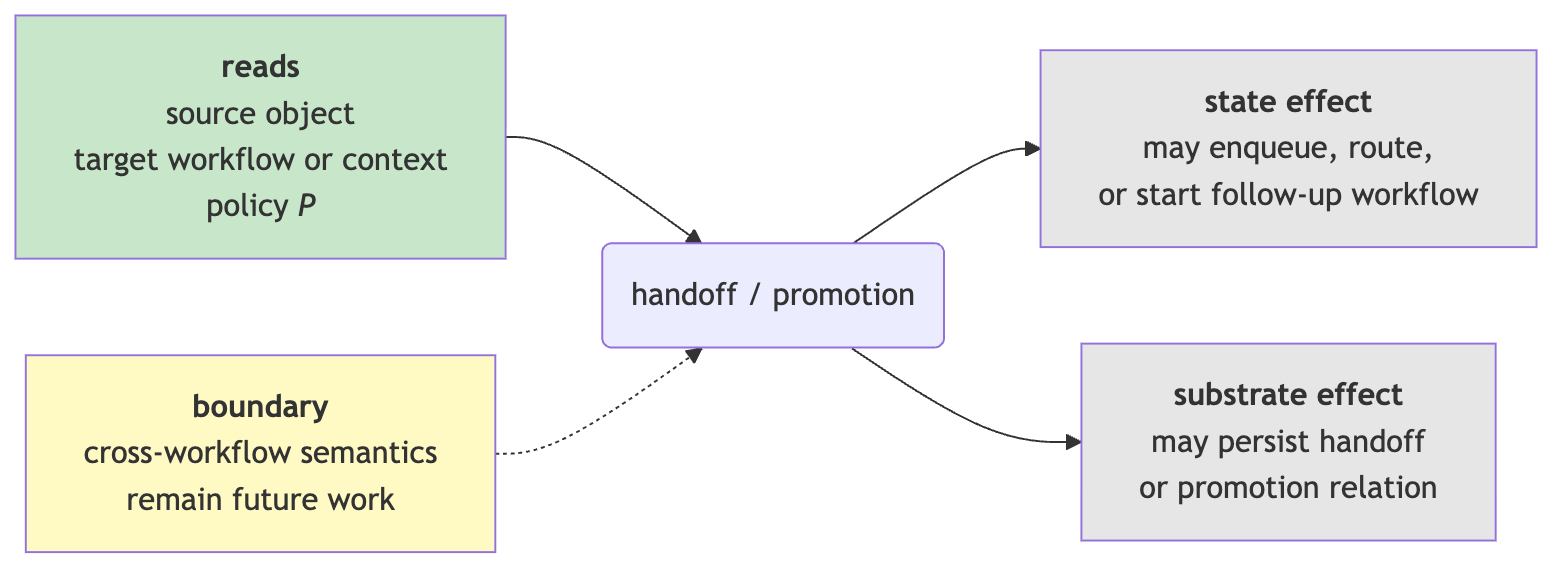}}
\end{center}

\section{Appendix B. Worked Example: ARS-Style Claim Review
Workflow}\label{appendix-b.-worked-example-ars-style-claim-review-workflow}

This appendix gives one compact example of how the primitive vocabulary
from Appendix A can describe a contemporary academic-agent workflow
pattern. The example is inspired by public academic-research skill
systems such as Academic Research Skills for Claude Code (ARS), which
includes reviewer agents, integrity gates, claim verification,
material-passport-style metadata, response traces, and
publication-oriented artifacts
(\citeproc{ref-academicResearchSkills2026}{Imbad0202 2026}). The example
is not a reproduction, implementation, or evaluation of ARS. It
abstracts one claim-review pattern into the proposed object vocabulary.

The workflow reviews a single draft claim. It receives a claim from a
manuscript, assembles a bounded review context from permitted sources
and project records, checks deterministic preconditions, asks for
model-mediated claim-risk classification when appropriate, obtains an
author decision or reviewer-panel decision, and records the outcome. Its
purpose is to show how a workflow definition, workflow instance,
model-mediated inference, human authorization, structured deliberation,
and durable records can inhabit the same semantic object model.

\subsection{B.1 Worked Example
Overview}\label{b.1-worked-example-overview}

The proposed model represents the claim-review steps explicitly, as
declared workflow structure and typed records. A prompt-based academic
workflow harness may otherwise distribute the same semantics across
skill instructions, reviewer-agent prompts, citation reports,
integrity-gate logs, file metadata, and author response notes. The
translation below makes those latent objects explicit without claiming
that this is how any specific system is implemented internally.

The workflow uses the following primitives from Appendix A.

{\def\LTcaptype{none} % do not increment counter
\begin{longtable}[]{@{}
  >{\raggedright\arraybackslash}p{(\linewidth - 2\tabcolsep) * \real{0.1840}}
  >{\raggedright\arraybackslash}p{(\linewidth - 2\tabcolsep) * \real{0.4800}}@{}}
\toprule\noalign{}
\begin{minipage}[b]{\linewidth}\raggedright
Primitive
\end{minipage} & \begin{minipage}[b]{\linewidth}\raggedright
Role in the example
\end{minipage} \\
\midrule\noalign{}
\endhead
\bottomrule\noalign{}
\endlastfoot
\texttt{resource} & Declares the manuscript section, checked source
notes, citation anchors, feedback records, and provenance-like material
available to the workflow. \\
\texttt{context} & Builds the bounded material visible to the
claim-review step. \\
\texttt{guard} & Prevents review from proceeding when required material
is absent or when the requested operation exceeds policy. \\
\texttt{derive} & Computes deterministic checks over claim shape,
required fields, and available source material. \\
\texttt{infer} & Requests a model-mediated claim-risk or claim-support
classification. \\
\texttt{approval} & Lets the author accept, reject, revise, or defer the
proposed classification. \\
\texttt{panel} & Captures structured deliberation when a classification
requires reviewer-style discussion rather than simple authorization. \\
\texttt{record} & Persists the classification, author decision, panel
outcome, and dependency links. \\
\texttt{state} & Holds active workflow bindings while the instance
runs. \\
\texttt{branch} & Selects the declared path after deterministic checks,
author approval, or panel outcome. \\
\end{longtable}
}

The ARS-style feature being abstracted is not a full research pipeline.
It is a narrow claim-review slice: a claim is checked against available
material, classified for support/risk, optionally escalated to a
reviewer-style panel, and stored with enough context to support later
inspection.

\subsection{B.2 Object View}\label{b.2-object-view}

Before execution, the workflow exists as a \texttt{workflow-definition}
object:

\begin{itemize}
\tightlist
\item
  declared input: one \texttt{draft-claim};
\item
  declared resources: manuscript section, checked research notes,
  citation anchors, feedback records, and optional material-passport
  metadata;
\item
  declared state: current support classification, author decision, panel
  decision, and completion status;
\item
  declared steps: assemble context, check claim shape, classify
  support/risk, request author decision, optionally invoke panel review,
  and record the outcome;
\item
  declared policies: no direct model access to resources, no side
  effects from \texttt{infer}, bounded access to source material, and
  human approval before article-changing decisions.
\end{itemize}

When the executor starts the workflow, it creates a
\texttt{workflow-instance}. The instance owns active workflow state, but
the consequential outputs are represented as durable records in the
knowledge substrate.

\begin{samepage}
\begin{center}
\pandocbounded{\includegraphics[width=0.82\linewidth,height=0.62\textheight,keepaspectratio]{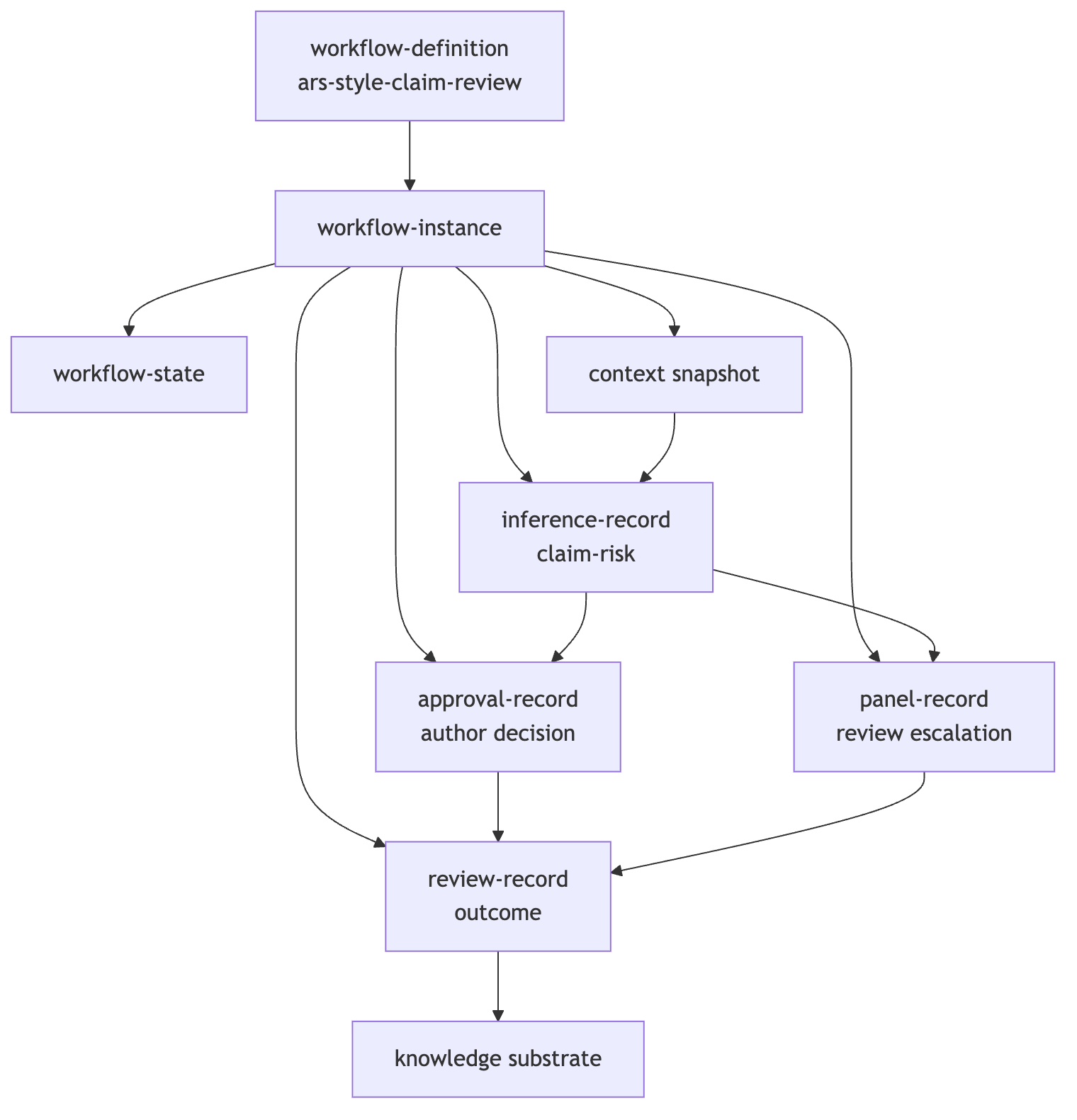}}
\end{center}

\ArticleCaption{Figure B1. An ARS-style claim-review workflow represented as a
definition, an instance, active state, and durable mediated records.}
\end{samepage}

\subsection{B.3 Pseudo-Lisp Workflow
Definition}\label{b.3-pseudo-lisp-workflow-definition}

The following sketch shows the workflow in a Lisp-like notation. The
forms denote semantic objects and executor-interpreted constructs; they
are not unrestricted executable code.

\begin{Shaded}
\begin{Highlighting}[]
\NormalTok{(define{-}workflow ars{-}style{-}claim{-}review}
  \BuiltInTok{:version} \StringTok{"1.0"}
\NormalTok{  :scope :single{-}claim}
\NormalTok{  :source{-}pattern :contemporary{-}academic{-}agent{-}workflow}

\NormalTok{  (inputs}
\NormalTok{    (input claim }\BuiltInTok{:type}\NormalTok{ draft{-}claim :required }\KeywordTok{t}\NormalTok{))}

\NormalTok{  (resources}
\NormalTok{    (resource manuscript{-}section}
\NormalTok{      :kind document{-}section}
\NormalTok{      :access :read)}
\NormalTok{    (resource checked{-}source{-}notes}
\NormalTok{      :kind research{-}notes}
\NormalTok{      :access :read)}
\NormalTok{    (resource citation{-}anchors}
\NormalTok{      :kind source{-}locators}
\NormalTok{      :access :read)}
\NormalTok{    (resource feedback{-}records}
\NormalTok{      :kind review{-}history}
\NormalTok{      :access :read)}
\NormalTok{    (resource material{-}passport}
\NormalTok{      :kind provenance{-}summary}
\NormalTok{      :access :read}
\NormalTok{      :required }\KeywordTok{nil}\NormalTok{))}

\NormalTok{  (session{-}state}
\NormalTok{    source{-}available?}
\NormalTok{    claim{-}shape{-}ok?}
\NormalTok{    claim{-}risk}
\NormalTok{    author{-}decision}
\NormalTok{    panel{-}decision}
\NormalTok{    completion{-}status)}

\NormalTok{  (rules}
\NormalTok{    (guard :source{-}material{-}required}
      \BuiltInTok{:predicate}\NormalTok{ (available? checked{-}source{-}notes)}
\NormalTok{      :on{-}fail (record :missing{-}source{-}material))}

\NormalTok{    (guard :citation{-}anchor{-}required}
      \BuiltInTok{:predicate}\NormalTok{ (has{-}locator? claim citation{-}anchors)}
\NormalTok{      :on{-}fail (record :anchorless{-}claim))}

\NormalTok{    (guard :no{-}model{-}side{-}effects}
\NormalTok{      :applies{-}to infer}
\NormalTok{      :policy }\BuiltInTok{:read{-}only}\NormalTok{))}

\NormalTok{  (steps}
\NormalTok{    (}\KeywordTok{step}\NormalTok{ :assemble{-}context}
\NormalTok{      (context claim{-}review{-}context}
\NormalTok{        :from (claim}
\NormalTok{               manuscript{-}section}
\NormalTok{               checked{-}source{-}notes}
\NormalTok{               citation{-}anchors}
\NormalTok{               feedback{-}records}
\NormalTok{               material{-}passport)))}

\NormalTok{    (}\KeywordTok{step}\NormalTok{ :check{-}claim{-}shape}
\NormalTok{      (bind claim{-}shape{-}ok?}
\NormalTok{        (derive :claim{-}shape{-}ok?}
\NormalTok{          :from claim}
\NormalTok{          :as (has{-}fields? claim :text :section :support))))}

\NormalTok{    (}\KeywordTok{step}\NormalTok{ :check{-}source{-}availability}
\NormalTok{      (bind source{-}available?}
\NormalTok{        (derive :source{-}available?}
\NormalTok{          :from (claim checked{-}source{-}notes citation{-}anchors)}
\NormalTok{          :as (}\KeywordTok{and}\NormalTok{ (available? checked{-}source{-}notes)}
\NormalTok{                   (has{-}locator? claim citation{-}anchors)))))}

\NormalTok{    (branch (}\KeywordTok{and}\NormalTok{ claim{-}shape{-}ok? source{-}available?)}
\NormalTok{      (:false}
\NormalTok{        (record unresolved{-}claim}
\NormalTok{          :for claim}
\NormalTok{          :reason :insufficient{-}review{-}material}
\NormalTok{          :depends{-}on (claim{-}review{-}context}
\NormalTok{                       claim{-}shape{-}ok?}
\NormalTok{                       source{-}available?)))}

\NormalTok{      (:true}
\NormalTok{        (}\KeywordTok{step}\NormalTok{ :classify{-}claim{-}risk}
\NormalTok{          (bind claim{-}risk}
\NormalTok{            (infer :classify{-}claim{-}risk}
\NormalTok{              :context claim{-}review{-}context}
\NormalTok{              :returns claim{-}risk{-}record}
\NormalTok{              :schema (:supported}
\NormalTok{                       :weakly{-}supported}
\NormalTok{                       :unsupported}
\NormalTok{                       :overstated}
\NormalTok{                       :needs{-}human{-}review)}
\NormalTok{              :policy :no{-}actions)))))}

\NormalTok{    (}\KeywordTok{step}\NormalTok{ :author{-}decision}
\NormalTok{      (bind author{-}decision}
\NormalTok{        (approval :claim{-}risk{-}decision}
\NormalTok{          :request claim{-}risk}
\NormalTok{          :context claim{-}review{-}context}
\NormalTok{          :options (:accept}
\NormalTok{                    :revise}
\NormalTok{                    :defer}
\NormalTok{                    :escalate{-}to{-}panel))))}

\NormalTok{    (branch author{-}decision}
\NormalTok{      (:accept}
\NormalTok{        (record accepted{-}claim{-}review}
\NormalTok{          :claim claim}
\NormalTok{          :classification claim{-}risk}
\NormalTok{          :depends{-}on (claim{-}review{-}context}
\NormalTok{                       claim{-}risk/inference{-}record}
\NormalTok{                       author{-}decision/approval{-}record)))}

\NormalTok{      (:revise}
\NormalTok{        (record revision{-}needed}
\NormalTok{          :claim claim}
\NormalTok{          :classification claim{-}risk}
\NormalTok{          :depends{-}on (claim{-}review{-}context}
\NormalTok{                       claim{-}risk/inference{-}record}
\NormalTok{                       author{-}decision/approval{-}record)))}

\NormalTok{      (:defer}
\NormalTok{        (record research{-}needed}
\NormalTok{          :claim claim}
\NormalTok{          :classification claim{-}risk}
\NormalTok{          :depends{-}on (claim{-}review{-}context}
\NormalTok{                       claim{-}risk/inference{-}record}
\NormalTok{                       author{-}decision/approval{-}record)))}

\NormalTok{      (:escalate{-}to{-}panel}
\NormalTok{        (}\KeywordTok{step}\NormalTok{ :review{-}panel}
\NormalTok{          (bind panel{-}decision}
\NormalTok{            (panel :claim{-}review{-}panel}
\NormalTok{              :motion }\StringTok{"How should this claim be treated?"}
\NormalTok{              :context claim{-}review{-}context}
\NormalTok{              :uses claim{-}risk/inference{-}record}
\NormalTok{              :options (:accept{-}current{-}classification}
\NormalTok{                        :revise{-}claim}
\NormalTok{                        :defer{-}for{-}source{-}work)}
\NormalTok{              :arguments ((:for claim{-}risk)}
\NormalTok{                          (:against missing{-}or{-}weak{-}source{-}support)}
\NormalTok{                          (:uncertainty source{-}locator{-}quality)}
\NormalTok{                          (:prior feedback{-}records))))))))}

\NormalTok{    (branch panel{-}decision}
\NormalTok{      (:accept{-}current{-}classification}
\NormalTok{        (record accepted{-}after{-}panel}
\NormalTok{          :claim claim}
\NormalTok{          :classification claim{-}risk}
\NormalTok{          :panel panel{-}decision}
\NormalTok{          :depends{-}on (claim{-}risk/inference{-}record}
\NormalTok{                       panel{-}decision/panel{-}record)))}

\NormalTok{      (:revise{-}claim}
\NormalTok{        (record revision{-}needed{-}after{-}panel}
\NormalTok{          :claim claim}
\NormalTok{          :classification claim{-}risk}
\NormalTok{          :panel panel{-}decision}
\NormalTok{          :depends{-}on (claim{-}risk/inference{-}record}
\NormalTok{                       panel{-}decision/panel{-}record)))}

\NormalTok{      (:defer{-}for{-}source{-}work}
\NormalTok{        (record source{-}work{-}needed}
\NormalTok{          :claim claim}
\NormalTok{          :classification claim{-}risk}
\NormalTok{          :panel panel{-}decision}
\NormalTok{          :depends{-}on (claim{-}risk/inference{-}record}
\NormalTok{                       panel{-}decision/panel{-}record))))))}
\end{Highlighting}
\end{Shaded}

In this illustrative notation, paths such as
\texttt{claim-risk/inference-record},
\texttt{author-decision/approval-record}, and
\texttt{panel-decision/panel-record} denote records associated with
earlier bindings. They are dependency references, not new primitives.

The \texttt{panel} form still has no direct transition authority. It
creates a deliberative record that the executor may later use in a
declared branch. This preserves the paper's distinction between a
model-mediated classification, an author authorization, a structured
deliberation, and an executor-applied transition.

\subsection{B.4 Execution View}\label{b.4-execution-view}

The executor interprets the workflow definition. It resolves declared
resources, assembles context, evaluates deterministic expressions,
mediates model calls, applies policy, asks the author or panel for
decisions, and persists consequential records.

{\def\LTcaptype{none} % do not increment counter
\begin{longtable}[]{@{}
  >{\raggedright\arraybackslash}p{(\linewidth - 4\tabcolsep) * \real{0.3450}}
  >{\raggedright\arraybackslash}p{(\linewidth - 4\tabcolsep) * \real{0.1610}}
  >{\raggedright\arraybackslash}p{(\linewidth - 4\tabcolsep) * \real{0.4140}}@{}}
\toprule\noalign{}
\begin{minipage}[b]{\linewidth}\raggedright
Step
\end{minipage} & \begin{minipage}[b]{\linewidth}\raggedright
Primitive(s)
\end{minipage} & \begin{minipage}[b]{\linewidth}\raggedright
Main effect
\end{minipage} \\
\midrule\noalign{}
\endhead
\bottomrule\noalign{}
\endlastfoot
Resolve declared material & \texttt{resource} & The executor loads
permitted manuscript, source, citation-anchor, and feedback material;
the model does not fetch it directly. \\
Assemble review context & \texttt{context} & A bounded
\texttt{context-snapshot} is created for later review. \\
Check claim shape & \texttt{derive} & A deterministic expression
computes whether the claim has required fields. \\
Check source availability & \texttt{derive}, \texttt{guard} & The
executor verifies that required source notes and citation anchors
exist. \\
Stop or continue & \texttt{branch}, \texttt{record} & Missing required
material creates an unresolved-claim record and stops the review
path. \\
Classify claim risk & \texttt{infer} & The model proposes a support/\allowbreak{}risk
classification; the executor validates and records it. \\
Ask the author & \texttt{approval} & The author accepts, revises,
defers, or escalates the proposed classification. \\
Review disputed classification & \texttt{panel} & A structured
deliberation records motion, options, arguments, context, and dependency
on the inference record. \\
Persist outcome & \texttt{record} & The final decision is stored with
links to the claim, context snapshot, inference record, approval record,
and optional panel record. \\
\end{longtable}
}

\begin{samepage}
\begin{center}\resizebox{!}{0.62\textheight}{\begin{tikzpicture}[
  font=\sffamily\small,
  >=Latex,
  node distance=10mm and 16mm,
  line/.style={-{Latex[length=2.4mm,width=1.8mm]}, draw=slate, line width=0.75pt},
  branch/.style={line, font=\sffamily\scriptsize, text=slate},
  process/.style={
    rectangle,
    rounded corners=2.5pt,
    draw=bluegray,
    fill=bluegray!8,
    line width=0.8pt,
    align=center,
    text width=36mm,
    minimum height=9mm,
    inner xsep=3mm,
    inner ysep=2mm
  },
  decision/.style={
    diamond,
    aspect=2.0,
    draw=amber!70!black,
    fill=amber!12,
    line width=0.8pt,
    align=center,
    text width=28mm,
    inner sep=1.2mm
  },
  record/.style={
    process,
    draw=green!45!black,
    fill=green!8
  },
  panel/.style={
    process,
    draw=purple!60!black,
    fill=purple!7
  }
]

% Color definitions
\definecolor{bluegray}{RGB}{79,105,132}
\definecolor{slate}{RGB}{52,63,76}
\definecolor{amber}{RGB}{214,145,38}

% Nodes
\node[process] (A) {input draft-claim};
\node[process, below=of A] (B) {resource\\resolve manuscript, sources, anchors};
\node[process, below=of B] (C) {context\\assemble claim-review-context};

\node[process, above right=of C] (D) {derive\\claim-shape-ok?};
\node[process, below right=of C] (E) {derive\\source-available?};
\node[decision, below right=of D] (F) {shape and source ok?};

\node[record, above right=of F] (G) {record\\unresolved-claim};
\node[process, below right=of F] (H) {infer\\classify-claim-risk};
\node[process, below=of H] (I) {approval\\author-decision};
\node[decision, below=of I] (J) {decision};

\node[record, above left= of J] (K) {record\\accepted-claim-review};
\node[record, left=of J] (L) {record\\revision-needed};
\node[record, below left= of J] (M) {record\\research-needed};
\node[panel, below= of J] (N) {panel\\claim-review-panel};

\node[decision, below=of N] (O) {panel outcome};
\node[record, below left=18mm and 28mm of O] (P) {record\\accepted-after-panel};
\node[record, below=18mm of O] (Q) {record\\revision-needed-after-panel};
\node[record, below right=18mm and 28mm of O] (R) {record\\source-work-needed};

% Edges
\draw[line] (A) -- (B);
\draw[line] (B) -- (C);
\draw[line] (C) -- (D);
\draw[line] (C) -- (E);
\draw[line] (D) -- (F);
\draw[line] (E) -- (F);

\draw[branch] (F) -- node[above left, pos=0.45] {no} (G);
\draw[branch] (F) -- node[above right, pos=0.45] {yes} (H);
\draw[line] (H) -- (I);
\draw[line] (I) -- (J);

\draw[branch] (J) -- node[above left, pos=0.42] {accept} (K);
\draw[branch] (J) -- node[right, pos=0.38] {revise} (L);
\draw[branch] (J) -- node[above right, pos=0.42] {defer} (M);
\draw[branch] (J) -- node[above, pos=0.48] {escalate} (N);

\draw[line] (N) -- (O);
\draw[branch] (O) -- node[above left, pos=0.42] {accept} (P);
\draw[branch] (O) -- node[right, pos=0.38] {revise} (Q);
\draw[branch] (O) -- node[above right, pos=0.42] {defer} (R);

\end{tikzpicture}}\end{center}

\ArticleCaption{Figure B2. A compact execution view of an ARS-style claim-review
workflow, including deterministic checks, model-mediated classification,
author approval, and reviewer-style panel escalation.}
\end{samepage}

\subsection{B.5 State And Records}\label{b.5-state-and-records}

The workflow instance may keep active state while it runs:

\begin{Shaded}
\begin{Highlighting}[]
\NormalTok{(:state}
\NormalTok{  (source{-}available? }\KeywordTok{t}\NormalTok{)}
\NormalTok{  (claim{-}shape{-}ok? }\KeywordTok{t}\NormalTok{)}
\NormalTok{  (claim{-}risk :weakly{-}supported)}
\NormalTok{  (author{-}decision :escalate{-}to{-}panel)}
\NormalTok{  (panel{-}decision :defer{-}for{-}source{-}work)}
\NormalTok{  (completion{-}status :source{-}work{-}needed))}
\end{Highlighting}
\end{Shaded}

That state is not the same as the durable knowledge objects. The durable
part is the record structure created around the run:

\begin{Shaded}
\begin{Highlighting}[]
\NormalTok{(record source{-}work{-}needed}
\NormalTok{  :claim claim}
\NormalTok{  :classification claim{-}risk}
\NormalTok{  :context claim{-}review{-}context}
\NormalTok{  :inference claim{-}risk/inference{-}record}
\NormalTok{  :approval author{-}decision/approval{-}record}
\NormalTok{  :panel panel{-}decision/panel{-}record}
\NormalTok{  :status :open)}
\end{Highlighting}
\end{Shaded}

A panel produces a deliberative record rather than a simple
authorization record:

\begin{Shaded}
\begin{Highlighting}[]
\NormalTok{(panel{-}record claim{-}review{-}panel}
\NormalTok{  :motion }\StringTok{"How should this claim be treated?"}
\NormalTok{  :options (:accept{-}current{-}classification}
\NormalTok{            :revise{-}claim}
\NormalTok{            :defer{-}for{-}source{-}work)}
\NormalTok{  :arguments ((:for claim{-}risk)}
\NormalTok{              (:against missing{-}or{-}weak{-}source{-}support)}
\NormalTok{              (:uncertainty source{-}locator{-}quality)}
\NormalTok{              (:prior feedback{-}records))}
\NormalTok{  :context claim{-}review{-}context}
\NormalTok{  :depends{-}on (claim{-}risk/inference{-}record)}
\NormalTok{  :decision :defer{-}for{-}source{-}work)}
\end{Highlighting}
\end{Shaded}

The distinction is important. Active state supports execution and
resumption. Records support later inspection, comparison, review, and
possible reuse as context for another workflow. The panel record also
preserves a structured deliberation path without treating it as a proof
of correctness, source support, or audit quality.

\subsection{B.6 Optional Batch Form}\label{b.6-optional-batch-form}

The same definition can be lifted into a bounded loop when several
claims must be reviewed. The loop is still a declared DSL construct, not
hidden control flow in an external script.

\begin{Shaded}
\begin{Highlighting}[]
\NormalTok{(}\KeywordTok{loop}\NormalTok{ :claims}
\NormalTok{  :over candidate{-}claims}
\NormalTok{  :until (derive :all{-}reviewed?)}

\NormalTok{  (}\KeywordTok{step}\NormalTok{ :review{-}one{-}claim}
\NormalTok{    (run{-}workflow ars{-}style{-}claim{-}review}
\NormalTok{      :claim current{-}claim)))}
\end{Highlighting}
\end{Shaded}

This batch form does not change the meaning of \texttt{infer},
\texttt{approval}, or \texttt{panel}. Each claim review still creates
its own context snapshot, inference record, approval record, optional
panel record, and final review record. The \texttt{run-workflow} denotes
the executor invoking a named workflow definition as a sub-workflow
instance under the current context and capability policy; its full
semantics, including record linkage between parent and child instances,
are part of the handoff/promotion refinement track in Future Work.

\section{Appendix C. Preliminary PROV-DM
Mapping}\label{appendix-c.-preliminary-prov-dm-mapping}

This appendix gives a preliminary comparison between the proposed
semantic object model and W3C PROV-DM / scientific workflow provenance
concepts. The mapping is comparative, not a formal PROV serialization.
It shows that the model is provenance-compatible but not
provenance-complete: provenance concepts can describe many entities,
activities, and relations after they occur, while the proposed model
defines DSL-level roles, executor mediation, LLM-specific inference
boundaries, panel structure, context policy, and lifecycle questions
that remain application semantics. Candidate refinements such as
\texttt{capability} / \texttt{action}, \texttt{handoff} /
\texttt{promotion}, and lifecycle relations such as
\texttt{supersession-link} are omitted from this preliminary table
because their side-effect, cross-workflow, and lifecycle semantics are
deferred to Future Work; mapping them here would be speculative.

\textbf{Operations with closer PROV analogues}

{\def\LTcaptype{none} % do not increment counter
\begin{longtable}[]{@{}
  >{\raggedright\arraybackslash}p{(\linewidth - 4\tabcolsep) * \real{0.2300}}
  >{\raggedright\arraybackslash}p{(\linewidth - 4\tabcolsep) * \real{0.3450}}
  >{\raggedright\arraybackslash}p{(\linewidth - 4\tabcolsep) * \real{0.3450}}@{}}
\toprule\noalign{}
\begin{minipage}[b]{\linewidth}\raggedright
Object
\end{minipage} & \begin{minipage}[b]{\linewidth}\raggedright
PROV analogue
\end{minipage} & \begin{minipage}[b]{\linewidth}\raggedright
What the model adds
\end{minipage} \\
\midrule\noalign{}
\endhead
\bottomrule\noalign{}
\endlastfoot
\texttt{workflow-definition} & PROV \texttt{Plan} or
workflow-specification entity & DSL primitives, state schema, guards,
context declarations, approvals, panels, and
\texttt{derive}/\allowbreak{}\texttt{infer} boundaries \\
\texttt{workflow-instance} & Workflow-run \texttt{Activity} plus
persisted checkpoint entity & Intentionally hybrid: active execution
occurrence and durable resumable object linking mediated records \\
\texttt{derive} & Activity with declared inputs generating a derived
entity & DSL-level commitment to deterministic, replayable computation
without LLM judgment \\
\texttt{infer} /\allowbreak{} \texttt{inference-record} & Activity/\allowbreak{}entity pair:
activity uses context, prompt, model, and schema & Executor validation,
capability policy, and the rule that model output is not transition
authority \\
\texttt{record} & Entity generation through an executor-mediated
activity & Semantic boundary where workflow material becomes part of the
knowledge substrate under lifecycle policy \\
\texttt{approval-record} & Decision or authorization entity generated by
a human or policy activity & Workflow-authority semantics: permits,
rejects, or defers a transition under explicit policy \\
\texttt{panel-record} & Deliberation activity generating a decision or
deferral entity & Motion, options, structured arguments, evidence links,
deferral, and distinction from simple authorization; Appendix B gives a
trace \\
\texttt{context-snapshot} & Entity, collection, or bundle used by an
inference or approval activity & Visibility policy, source authority,
progressive disclosure, freshness, and review-scope semantics \\
\texttt{dependency-link} & \texttt{used}, \texttt{wasDerivedFrom},
\texttt{wasInformedBy}, or \texttt{wasInfluencedBy} & Typed dependency
roles: premise, branch condition, validation evidence, model input, or
review basis \\
\end{longtable}
}

\textbf{Operations without direct PROV analogues}

{\def\LTcaptype{none} % do not increment counter
\begin{longtable}[]{@{}
  >{\raggedright\arraybackslash}p{(\linewidth - 4\tabcolsep) * \real{0.2509}}
  >{\raggedright\arraybackslash}p{(\linewidth - 4\tabcolsep) * \real{0.3345}}
  >{\raggedright\arraybackslash}p{(\linewidth - 4\tabcolsep) * \real{0.3345}}@{}}
\toprule\noalign{}
\begin{minipage}[b]{\linewidth}\raggedright
Object
\end{minipage} & \begin{minipage}[b]{\linewidth}\raggedright
PROV analogue
\end{minipage} & \begin{minipage}[b]{\linewidth}\raggedright
What the model adds
\end{minipage} \\
\midrule\noalign{}
\endhead
\bottomrule\noalign{}
\endlastfoot
\texttt{state} & Persisted state snapshot, when recorded & Active
workflow state is not durable evidence until \texttt{record} or a
snapshot promotes it into the knowledge substrate \\
\texttt{guard} & Policy or constraint activity influencing a transition
& The executor enforces permission, scope, violation handling, and
transition denial; PROV records that a guard existed \\
\texttt{branch} /\allowbreak{} control transition & Control-flow influence or
informed activity & A branch remains executor-applied even when its
condition depends on \texttt{infer}; influence is recorded without
giving the LLM control-flow authority \\
\end{longtable}
}

The proposed object model does not replace PROV-DM, and the table does
not claim audit quality, trust, accountability, attribution,
reproducibility, or provenance correctness. Formal PROV export,
lifecycle semantics, and governance evaluation remain Future Work.

\section{Appendix D. Exploratory Skill-Scan
Notes}\label{appendix-d.-exploratory-skill-scan-notes}

Appendix D supports Section 4 by reporting the exploratory scan's corpus
grouping, primitive-count table, and source-use cautions. The scan
covered 77 skill-like workflow artifacts. The corpus was selected rather
than random, and the scoring was qualitative. The scan should therefore
be read as vocabulary exploration, not as empirical validation of the
thesis.

The materials were grouped into three broad categories:

{\def\LTcaptype{none} % do not increment counter
\begin{longtable}[]{@{}
  >{\raggedright\arraybackslash}p{(\linewidth - 6\tabcolsep) * \real{0.2800}}
  >{\raggedleft\arraybackslash}p{(\linewidth - 6\tabcolsep) * \real{0.0800}}
  >{\raggedright\arraybackslash}p{(\linewidth - 6\tabcolsep) * \real{0.2800}}
  >{\raggedright\arraybackslash}p{(\linewidth - 6\tabcolsep) * \real{0.2800}}@{}}
\toprule\noalign{}
\begin{minipage}[b]{\linewidth}\raggedright
Group
\end{minipage} & \begin{minipage}[b]{\linewidth}\raggedleft
Items
\end{minipage} & \begin{minipage}[b]{\linewidth}\raggedright
Evidence status
\end{minipage} & \begin{minipage}[b]{\linewidth}\raggedright
Use in this paper
\end{minipage} \\
\midrule\noalign{}
\endhead
\bottomrule\noalign{}
\endlastfoot
Local/\allowbreak{}internal design material & 17 & Internal structural evidence only
& Folded into design motivation, not used as public evidence \\
Codex/\allowbreak{}Claude-style agent artifacts & 30 & Exploratory public evidence &
Primitive refinement \\
Workflow/\allowbreak{}OpenClaw operational artifacts & 30 & Ecosystem-specific
exploratory evidence & State, action, recovery, and record patterns \\
Total & 77 & Selected qualitative corpus & Vocabulary motivation only \\
\end{longtable}
}

The following table reports, for each primitive, how many items in each
group scored \texttt{2} or \texttt{3} on the original 0-3 checklist.
These counts describe the pilot corpus only.

{\def\LTcaptype{none} % do not increment counter
\begin{longtable}[]{@{}
  >{\raggedright\arraybackslash}p{(\linewidth - 6\tabcolsep) * \real{0.4800}}
  >{\raggedleft\arraybackslash}p{(\linewidth - 6\tabcolsep) * \real{0.1360}}
  >{\raggedleft\arraybackslash}p{(\linewidth - 6\tabcolsep) * \real{0.1400}}
  >{\raggedleft\arraybackslash}p{(\linewidth - 6\tabcolsep) * \real{0.1400}}@{}}
\toprule\noalign{}
\begin{minipage}[b]{\linewidth}\raggedright
Primitive
\end{minipage} & \begin{minipage}[b]{\linewidth}\raggedleft
Local/\allowbreak{}internal (17)
\end{minipage} & \begin{minipage}[b]{\linewidth}\raggedleft
Codex/\allowbreak{}Claude-style (30)
\end{minipage} & \begin{minipage}[b]{\linewidth}\raggedleft
Workflow/\allowbreak{}OpenClaw (30)
\end{minipage} \\
\midrule\noalign{}
\endhead
\bottomrule\noalign{}
\endlastfoot
\texttt{input} & 15 & 18 & 29 \\
\texttt{resource} & 17 & 24 & 30 \\
\texttt{derive} & 16 & 25 & 30 \\
\texttt{infer} & 16 & 20 & 26 \\
\texttt{guard} & 17 & 26 & 30 \\
\texttt{state} & 11 & 11 & 27 \\
\texttt{step} & 17 & 26 & 30 \\
\texttt{loop} & 11 & 10 & 27 \\
\texttt{branch} & 17 & 21 & 30 \\
\texttt{panel} & 8 & 3 & 9 \\
\texttt{context} & 16 & 26 & 30 \\
\texttt{tool} & 6 & 14 & 26 \\
\texttt{record} & 13 & 21 & 30 \\
\texttt{resume} & 0 & 3 & 16 \\
\texttt{serialize} & 9 & 5 & 17 \\
\end{longtable}
}

Several cautions follow from the scoring process.

First, \texttt{approval} was not part of the original checklist. It
emerged during synthesis because many artifacts contained permission or
confirmation gates that were not full deliberative panels. The table
therefore reports \texttt{panel} scores only; it does not provide
numeric approval counts.

Second, \texttt{context} was overloaded. The scan encountered
source-authority context, inference-visible context, operating context,
and progressive-disclosure context. These should be separated or
annotated in future scoring.

Third, \texttt{state} and \texttt{record} overlap in many file-backed
workflows. A task file, log, checkpoint, or memory entry can be active
execution state during a run and a durable record after the run.

Fourth, local/internal design material is not public evidence. It
remains useful because it exposed recurring patterns such as approval,
promotion, handoff, and review queues, but examples from that group
should not be used as public claims without separate sanitization.

Finally, the operational artifacts scored higher on \texttt{state},
\texttt{tool}, \texttt{record}, \texttt{resume}, and \texttt{serialize}
than the Codex/Claude-style group. This suggests that side-effecting
capabilities, recovery behavior, and persistent operating memory may
require explicit modeling in future versions of the vocabulary.

\section{References}\label{references}

\protect\phantomsection\label{refs}

\end{document}